\documentclass[pdflatex,sn-mathphys-num, iicol]{sn-jnl}

\usepackage{graphicx}%
\usepackage{multirow}%
\usepackage{amsmath,amssymb,amsfonts}%
\usepackage{amsthm}%
\usepackage{mathrsfs}%
\usepackage[title]{appendix}%
\usepackage{textcomp}%
\usepackage{manyfoot}%
\usepackage{booktabs}%
\usepackage{algorithm}%
\usepackage{algorithmicx}%
\usepackage{algpseudocode}%
\usepackage{listings}%

\usepackage{makecell}
\usepackage{pifont}
\usepackage[table]{xcolor}
\usepackage{afterpage}
\usepackage{float}
\definecolor{DeepRed}{RGB}{184,42,42}
\definecolor{DeepBlue}{RGB}{58,108,158}

\theoremstyle{thmstyleone}%
\theoremstyle{thmstyletwo}%

\theoremstyle{thmstylethree}%

\begin{document}
\flushbottom

\title{AHDMIL: Asymmetric Hierarchical Distillation Multi-Instance Learning for Fast and Accurate Whole-Slide Image Classification}

\author[1]{\fnm{Jiuyang} \sur{Dong}}\email{jiuyang.dong@stu.hit.edu.cn}
\author[1]{\fnm{Jiahan} \sur{Li}}\email{jiahan.li@stu.hit.edu.cn}
\author*[1]{\fnm{Junjun} \sur{Jiang}}\email{jiangjunjun@hit.edu.cn}
\author[1]{\fnm{Kui} \sur{Jiang}}\email{jiangkui@hit.edu.cn}
\author*[1,2]{\fnm{Yongbing} \sur{Zhang}}\email{yzhang08@hit.edu.cn}

\affil*[1]{\orgdiv{Computer Science and Technology}, \orgname{Harbin Institute of Technology}, \orgaddress{\city{Harbin}, \postcode{150006}, \state{Heilongjiang}, \country{China}}}

\affil*[2]{\orgdiv{Computer Science and Technology}, \orgname{Harbin Institute of Technology (Shenzhen)}, \orgaddress{\city{Shenzhen}, \postcode{518000}, \state{Guangdong}, \country{China}}}


\abstract{
Although multi-instance learning (MIL) has succeeded in pathological image classification, it faces the challenge of high inference costs due to the need to process thousands of patches from each gigapixel whole slide image (WSI).
To address this, we propose AHDMIL, an Asymmetric Hierarchical Distillation Multi-Instance Learning framework that enables fast and accurate classification by eliminating irrelevant patches through a two-step training process.
AHDMIL comprises two key components: the Dynamic Multi-Instance Network (DMIN), which operates on high-resolution WSIs, and the Dual-Branch Lightweight Instance Pre-screening Network (DB-LIPN), which analyzes corresponding low-resolution counterparts.
In the first step, self-distillation (SD), DMIN is trained for WSI classification while generating per-instance attention scores to identify irrelevant patches.
These scores guide the second step, asymmetric distillation (AD), where DB-LIPN learns to predict the relevance of each low-resolution patch.
The relevant patches predicted by DB-LIPN have spatial correspondence with patches in high-resolution WSIs, which are used for fine-tuning and efficient inference of DMIN.
In addition, we design the first Chebyshev-polynomial-based Kolmogorov-Arnold (CKA) classifier in computational pathology, which improves classification performance through learnable activation layers.
Extensive experiments on four public datasets demonstrate that AHDMIL consistently outperforms previous state-of-the-art methods in both classification performance and inference speed. 
For example, on the Camelyon16 dataset, it achieves a relative improvement of 5.3\% in accuracy and accelerates inference by 1.2$\times$. Across all datasets, area under the curve (AUC), accuracy, f1 score, and brier score show consistent gains, with average inference speedups ranging from 1.2$\times$ to 2.1$\times$.
The code is available at https://github.com/JiuyangDong/AHDMIL.
}




\maketitle

\section{Introduction}\label{sec:intro}
\begin{figure}[!t]
\centering
\includegraphics[width=0.5\textwidth]{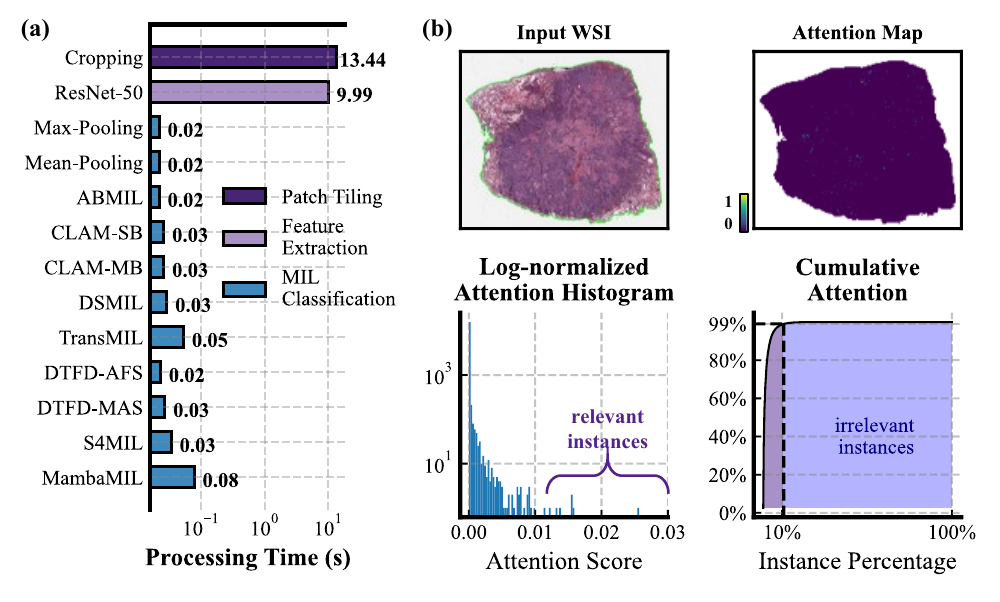}
\caption{\textbf{What makes inference slow?}
(a) Time-consuming data pre-processing: 
After comparing the time required for data pre-processing (patch tiling, feature extraction) and MIL network classification, it is clear that data pre-processing is the main speed bottleneck.
(b) Redundant irrelevant patches: 
For example, in a randomly selected WSI, numerous instances have extremely low attention scores, indicating their minimal contribution, if any, to the bag-level classification.}
\label{fig:teaser}
\end{figure}

Recently, multi-instance learning (MIL) has emerged as a leading paradigm for analyzing pathological whole slide images (WSIs), achieving notable success in tasks such as tumor detection and subtyping~\cite{lu2021data, li2021dual, zhang2022dtfd, shao2021transmil, fillioux2023structured, qu2022bi}, tissue micro-environment quantification~\cite{schapiro2017histocat, moen2019deep, mahmood2019deep, graham2019hover, saltz2018spatial, javed2020cellular}, and survival prediction~\cite{yang2024mambamil, shao2021weakly, chen2021whole, yao2020whole}. 
To enable these applications, MIL frameworks treat each gigapixel WSI as a bag consisting of thousands of patches, with each patch regarded as an instance.
Before classification, these patches must undergo computationally intensive tiling and feature extraction.
Taking the official Camelyon16~\cite{bejnordi2017diagnostic} test set as an example, Figure~\ref{fig:teaser} presents the average time per WSI spent on patch tiling, feature extraction, and inference across various MIL models.
As shown, data pre-processing dominates the processing time, taking orders of magnitude longer than the MIL inference itself.
Moreover, WSIs often contain a large number of redundant patches that contribute minimally to bag-level classification.
For example, aggregating the attention scores of just the top 10\% of patches already captures over 99\% of the total attention weight.
This suggests that a substantial portion of patches are redundant for classification and can be safely discarded without compromising model performance.

A straightforward way to reduce inference time is to discard irrelevant instances using attention scores.
However, existing MIL algorithms must extract features from all cropped patches before computing attention, resulting in a classic chicken-and-egg dilemma.
To address this, we propose an Asymmetric Hierarchical Distillation Multi-Instance Learning (AHDMIL) framework, which consists of two training steps, designed to efficiently identify irrelevant patches and enable fast yet accurate classification.
In the self-distillation (SD) step, instance-level features of all cropped patches from high-resolution WSIs are used to train the Dynamic Multi-Instance Network (DMIN).
The self-distillation constraint is introduced by enforcing prediction consistency between the teacher and student branches of DMIN, encouraging the teacher to focus on informative and task-relevant instances and enhancing the quality of instance selection.
The term ``self” implies that these two branches share the same architecture and network parameters, with the student learning from the teacher’s predictions on the full set of instances but receiving as input only a subset of instances assigned high attention scores by the teacher.
In the asymmetric distillation (AD) step, these attention scores are first used to guide the training of the Dual-Branch Lightweight Instance Pre-screening Network (DB-LIPN), which learns to predict patch relevance from low-resolution WSIs. 
The low-resolution regions identified as important by DB-LIPN correspond spatially to their high-resolution counterparts. 
Subsequently, these high-resolution patches are used to fine-tune the student branch of DMIN. 
During inference, the same DB-LIPN based selection enables skipping irrelevant high-resolution regions, thus reducing data pre-processing computational cost.
To further improve classification accuracy, we introduce a Chebyshev-polynomial-based Kolmogorov–Arnold (CKA) classifier with learnable activation layers. Unlike conventional linear classifiers, CKA significantly enhances nonlinear representation and decision-making capacity with negligible increase in parameters.

We summarize the primary contributions of this work as follows:
\begin{itemize}
    \item 
    This paper offers a crucial insight: eliminating irrelevant instances not only speeds up the inference process but also improves the classification performance. 
    This finding challenges the conventional trade-off between speed and performance and provides valuable inspiration for future research in multi-instance learning.
    \item We propose the DMIN for generating high-quality instance attention via a self-distillation strategy, the DB-LIPN for efficient patch selection to accelerate inference, and the CKA classifier for improved nonlinear representation with negligible overhead.
    \item Extensive experiments on four datasets demonstrate the effectiveness of our method. 
    For example, on the Camelyon16 dataset, AHDMIL achieves an area under the curve (AUC) of 91.96\% , an accuracy of 89.92\%, and an f1 score of 85.23\%, yielding relative gains of 4.8\%, 5.3\%, and 5.9\%,  respectively, over the previous best methods. Moreover, it consistently achieves at least a 1.2$\times$ inference speedup across all evaluated datasets.
\end{itemize}

A preliminary version of this work was presented in our earlier conference paper~\cite{dong2025fast}.
This extended journal version incorporates several significant improvements: 
1) We revise the second-step training strategy by replacing the sole distillation of LIPN from DMIN with a joint training of LIPN and DMIN, providing the potential for enhanced performance when combined with subsequent improvements.
2) We redesign the LIPN module and introduce DB-LIPN, a new dual-branch ensemble architecture aimed at mitigating overfitting to the validation set discussed in the conference version. To further stabilize training, we incorporate a cross-branch exponential moving average to reduce parameter divergence across branches in DB-LIPN.
3) Beyond the original binary classification tasks on the Camelyon16, TCGA-NSCLC, and TCGA-BRCA datasets, we extend our evaluation to a three-class subtype classification on the TCGA-RCC dataset.
4) We conduct more comprehensive experiments and analyses to strengthen empirical validation.
5) Our method achieves significantly better classification performance while further reducing inference time compared to the previous version.
Overall, these improvements substantially enhance the methodological rigor, empirical breadth, and practical impact of the original work, making this journal version a comprehensive and significant  contribution.

\begin{figure*}[!t]
    \centering
    \includegraphics[width=1.0\textwidth]{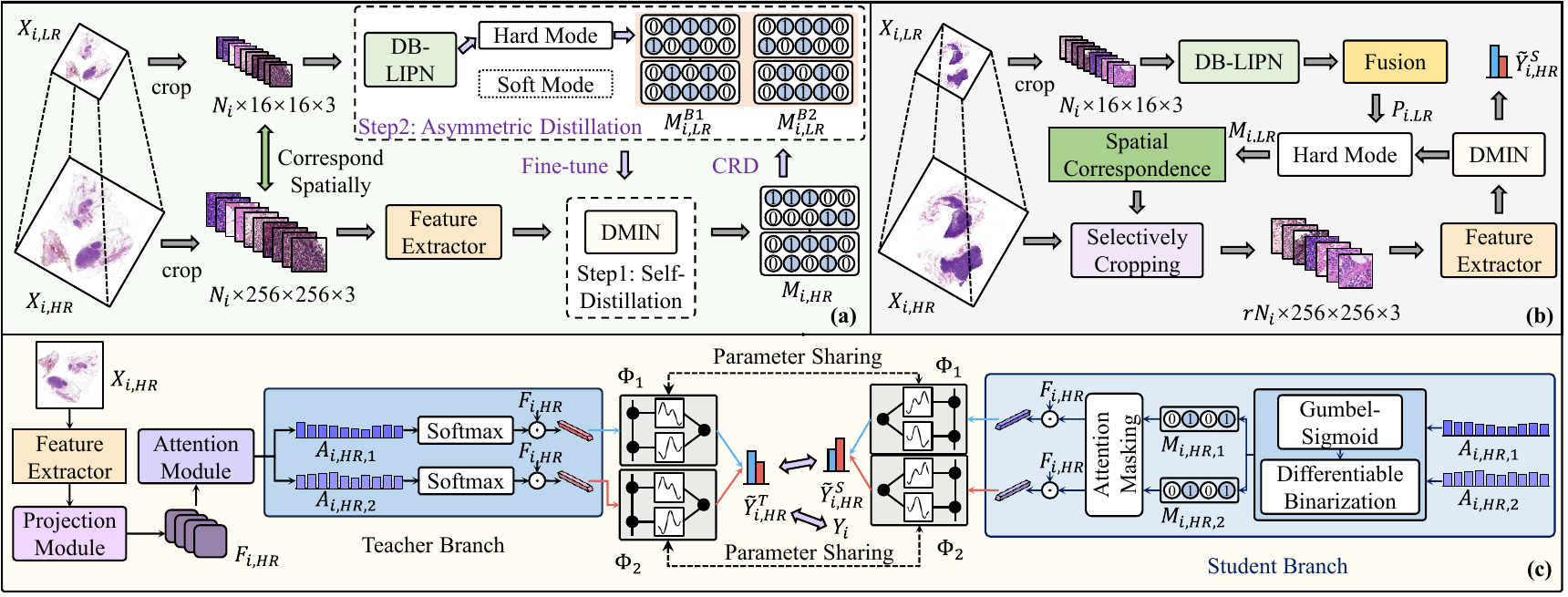}
    \caption{
    Overview of our AHDMIL framework. The framework is illustrated using a binary classification task as an example but is readily extendable to multi-class settings.
    (a) During training, DMIN is first trained on the high-resolution WSI $X_{i,HR}$ using a self-distillation strategy to generate per-instance attention scores. These scores then guide asymmetric distillation (AD) between DMIN and DB-LIPN, including both the cross-resolution distillation (CRD) of DB-LIPN and the fine-tuning of DMIN.
    One representative CRD strategy (`Hard Mode") is illustrated in the figure; additional variants are discussed in the main text.
    (b) During inference, DB-LIPN identifies classification-relevant regions in $X_{i,HR}$ by evaluating the low-resolution WSI $X_{i,LR}$.
    (c) Illustration of the self-distillation training of DMIN on the high-resolution input $X_{i,HR}$.
    }
    \label{fig:framework}
\end{figure*}

\section{Related Work}\label{sec:related}

In this section, we first provide a detailed review of recent advances in MIL, which forms the core of our work. 
Then, we briefly introduce dynamic neural networks and Kolmogorov–Arnold networks, highlighting their main ideas and relevance to our approach.

\subsection{MIL for WSI Classification}
MIL for WSI classification can be divided into instance-based and embedding-based approaches. 
Instance-based methods~\cite{zhu2017deep, pinheiro2015image, feng2017deep, kraus2016classifying, maron1997framework, keeler1990integrated, ramon2000multi} first classify each instance and then aggregate the predictions using Max-Pooling, Mean-Pooling, or other predefined pooling operations to generate the final bag-level prediction. 
In contrast, embedding-based methods first compute a bag-level representation by aggregating features from all instances, and then perform WSI classification.
ABMIL~\cite{ilse2018attention} employs an attention network to assess the significance of each instance and weight all instances accordingly. 
CLAM~\cite{lu2021data} extends ABMIL with either class-shared (CLAM-SB) or class-specific (CLAM-MB) attention branches, and introduces a clustering loss to regularize the instance feature space. 
DSMIL~\cite{li2021dual} selects the top-scoring instance from the instance-based branch and computes its similarity with other instances. These similarity scores are then used as attention weights to aggregate the bag-level representation in the embedding-based branch.
TransMIL~\cite{shao2021transmil} attempts to improve WSI representation by modeling  relationships among instances via self-attention.
DTFD-MIL~\cite{zhang2022dtfd} introduces pseudo-bags to augment training and employs a two-tier MIL framework to mitigate the label noise introduced by pseudo-bag partition.
S4MIL~\cite{fillioux2023structured} and MambaMIL~\cite{yang2024mambamil} adopt the state space model~\cite{gu2022parameterization} and Mamba architecture~\cite{gu2023mamba}, respectively, to better capture sequential dependencies among instances.

\subsection{Dynamic Neural Networks}
Dynamic neural networks~\cite{bolukbasi2017adaptive, teerapittayanon2016branchynet} adapt their architectures based on input data, enabling flexible control of computational redundancy.
In the era of visual transformers, many studies~\cite{zhao2024dynamic, liang2022not, meng2022adavit, rao2021dynamicvit, song2021dynamic, wang2021not} have improved inference efficiency by dynamically pruning redundant tokens.
This principle, suppressing uninformative elements while retaining essential semantics, lends itself naturally to multi-instance learning, where each bag may contain numerous redundant instances.
However, to our knowledge, this connection has not been explored. 
Motivated by this insight, we introduce dynamic network design into computational pathology for the first time, enabling instance-level redundancy reduction and efficient inference, thereby filling a methodological gap.
Moreover, our proposed AHDMIL framework addresses the aforementioned chicken-or-egg problem,  which is difficult to resolve by relying solely on end-to-end training, as is the case for many existing dynamic networks.

\subsection{Kolmogorov-Arnold Networks}
Most studies~\cite{sprecher2002space, koppen2002training, lin1993realization, lai2021kolmogorov, fakhoury2022exsplinet} prior to KAN~\cite{liu2024kan} explored the possibility of constructing neural networks based on the Kolmogorov-Arnold representation theore~\cite{schmidt2021kolmogorov} using the original two-layer structure.
Liu et al.~\cite{liu2024kan} extended this theorem to networks of arbitrary width and depth, investigating its potential as a foundational model for ``AI+Science.”
Subsequent research has mainly focused on either improving the integration of KAN into various tasks~\cite{knottenbelt2024coxkan, genet2024temporal, vaca2024kolmogorov, genet2024tkan, xu2024effective} or modifying its architecture~\cite{xu2024fourierkan, aghaei2024rkan, bodner2024convolutional, yang2024kolmogorov, bozorgasl2024wav, ta2024bsrbf}.
In this paper, we propose replacing the spline functions in KAN with first-kind Chebyshev polynomials, leading to a more expressive and efficient MIL classifier that overcomes KAN’s limitations in high-dimensional settings, thus improving classification performance on real-world pathological images.

\section{Method}\label{sec:method}

As illustrated in Figure~\ref{fig:framework}(a) and~(b), our proposed AHDMIL framework consists of two stages: training and inference.
The training stage comprises two steps, SD and AD, described in detail in Section~\ref{sec:SD} and~\ref{sec:AD}, respectively.
We first apply a self-distillation strategy to train DMIN on high-resolution WSIs, enabling it to perform bag-level classification and identify regions irrelevant to the classification task.
We then leverage the attention scores from the trained DMIN to guide cross-resolution distillation (CRD) for optimizing the DB-LIPN, which operates on low-resolution WSIs.
This allows DB-LIPN to assess the importance of each region with significantly reduced computational cost; the high-resolution patches deemed important, obtained via spatial correspondence, are then used to fine-tune the student branch of DMIN.
Similarly, during inference, DB-LIPN analyzes low-resolution WSIs to efficiently identify and discard regions irrelevant to classification.
The necessary high-resolution patches are then processed by a feature extractor and passed to DMIN to generate the final classification result.

It is worth noting that while both DMIN and DB-LIPN adopt dual-branch architectures, their designs differ substantially.
In DMIN, the teacher and student branches share the same network and parameters but perform distinct roles. 
In contrast, the two branches in DB-LIPN have identical architectures and are trained independently on the same data, with cross-branch exponential moving average (CBEMA) strategy used in some configurations to update one branch from the other.
Further implementation details are provided in the corresponding subsections.

\subsection{Data Preparation}
Before training, we preprocess the input data following the procedure proposed in CLAM~\cite{lu2021data}.
The dataset $\{X_{i}\}^{S}_{i=1}$ consists of $S$ WSI pyramids with slide-level labels, where each $X_i$ includes a pair of WSIs at high (20$\times$) and low (1.25$\times$) resolutions, respectively referred to as $X_{i,HR}$ and $X_{i,LR}$.
Although WSI pyramids typically include multiple magnification levels ranging from 1.25$\times$ to 40$\times$, this work focuses only on two representative magnifications.
After removing background regions, we extract $N_i$ pairs of corresponding patches: 16$\times$16 patches from $X_{i,LR}$ and 256$\times$256 patches from $X_{i,HR}$.
For details on how these paired patches are obtained, please refer to Section~\ref{sec:PRPT}.

\subsection{Self-Distillation Training Step}
\label{sec:SD}

As shown in Figure~\ref{fig:framework}(c), DMIN is designed to perform bag-level classification on high-resolution WSIs and simultaneously identify irrelevant instances.
It consists of five modules: the projection module, attention module, teacher branch, student branch, and CKA Classifier.

\subsubsection{Projection and Attention Modules}
During training, all patches tiled from the high-resolution WSI $X_{i,HR}$ are first passed through a pre-trained feature extractor to obtain a set of instance-level features $I_{i,HR}$.
These features are then fed into the projection module for dimensionality reduction, yielding a new feature set $F_{i,HR} \in \mathbb{R}^{N_i \times Q}$, where $Q$ denotes the dimensionality of the projected features. 
Next, $F_{i,HR}$ is passed to the attention module to compute instance-level attention scores:
\begin{equation}
    A_{i,HR} = [\phi({F_{i,HR}}V)\sigma({F_{i,HR}}U)]W,
\label{eq:attention}
\end{equation}
where $\phi(\cdot)$ and $\sigma(\cdot)$ denote the hyperbolic tangent and sigmoid functions, respectively.
Here, $U$, $V$, and $W$ are learnable weight matrices.
The attention module adopts the same architecture as that used in CLAM-MB~\cite{lu2021data}.
The resulting attention matrix for each class is denoted as $A_{i,HR,c} \in \mathbb{R}^{N_i \times 1}$, where $c \in \{1, 2, ..., C\}$, and $C$ represents  the total number of classes.

\subsubsection{Teacher Branch}
The dimension-reduced $F_{i,HR}$ is then linearly weighted by the attention matrix for each category to produce the bag-level representation, which is used for final classification:
\begin{equation}
    E_{i,HR,c}^{tea} = \varphi({A_{i,HR,c}})^{\top}{F_{i,HR,c}}, c\in\{1,2,...,C\}.
\label{eq:bag-level representation}
\end{equation}
Here, $\varphi(\cdot)$ denotes the softmax function and $E_{i,HR,c}^{tea}\in\mathbb{R}^{1\times Q}$ denotes the bag-level representation corresponding to the $c$-th category in the teacher branch.

\subsubsection{Student Branch} 
The student branch computes bag-level representations using only instances with relatively higher attention scores. 
We impose a constraint to promote consistency between these representations and those obtained by the teacher branch, which utilizes all instances.
This constraint, therefore, guides the attention module to focus more on instances that are critical for bag-level classification, while filtering out irrelevant ones.

However, directly selecting instances based on high attention scores involves a discrete operation, which leads to non-differentiability during optimization.
To address this, we employ the Gumbel trick~\cite{rao2021dynamicvit} to sample instances with high attention scores for end-to-end training.
First, we add Gumbel noise~\cite{herrmann2020channel} to the attention matrices and then apply the sigmoid function:
\begin{equation}
    \hat{A}_{i,HR,c} = \sigma(\frac{{A}_{i,HR,c}+G_{1,c}-G_{2,c}}{\tau}).
\label{eq:gumbel-sigmoid}
\end{equation}
Here, $G_{1,c} \in \mathbb{R}^{N_i \times 1}$ and $G_{2,c} \in \mathbb{R}^{N_i \times 1}$ are two noise matrices randomly sampled from the Gumbel distribution, and $\tau$ is the temperature parameter.
Next, we perform differentiable binarization of the sigmoid-transformed attention scores:
\begin{equation}
    {M}_{i,HR,c}^{j} = B(\hat{A}_{i,HR,c}^{j},\gamma)-D(\hat{A}_{i,HR,c}^{j})+\hat{A}_{i,HR,c}^{j},
\label{eq:discrete-mask}
\end{equation}
where ${M}_{i,HR,c}^{j}\in\{0,1\}$ denotes the mask value of the $j$-th instance, indicating whether the instance is selected (1) or masked out (0), and $\gamma$ is a threshold hyper-parameter.
$B(a,b)$ is a discrete binarization function that returns $1$ if $a > b$, and $0$ otherwise.
$D(\cdot)$ represents the gradient truncation operation.
Furthermore, we apply an attention masking mechanism to remove the influence of instances masked out (i.e., with zero mask values) on the bag-level representations:
\begin{equation}
    E_{i,HR,c}^{stu} = \sum_{j=1}^{N_i}\frac{exp({A}_{i,HR,c}^{j}){M}_{i,HR,c}^{j}}{\sum_{s=1}^{N_i}exp({A}_{i,HR,c}^{s}){M}_{i,HR,c}^{s}}F_{i,HR,c}^{j},
\label{eq:attention_mask}
\end{equation}
where $E_{i,HR,c}^{stu}\in \mathbb{R}^{1\times Q}$ denotes the bag-level representation of the $c$-th class.

\subsubsection{CKA Classifier} 
To enhance the classifier’s capacity, we adopt a KAN-style network that replaces fixed activations with nonlinear functions expressed as weighted sums of Chebyshev polynomial basis functions, benefiting from their stable polynomial approximation properties.
Specifically, we use the iterative form of $K$-degree Chebyshev polynomials to define the basis functions $T_{K}(x)$:
\begin{equation}
    T_{K}(x) = 2xT_{K-1}(x) - T_{K-2}(x), K\ge{2}.
\label{eq:chebyshev-polynomial}
\end{equation}
Here, $x\in\mathbb{R}^{1\times Q}$ represents the bag-level representation, with initial conditions $T_{0}(x)=\vec{\mathbf{1}}$ and $T_{1}(x)= x$. 
The CKA classifier output $\Phi(x)$ is obtained by linearly combining these basis functions $T(x)$ with learnable coefficients $\Omega\in\mathbb{R}^{Q\times(K+1)}$:
\begin{equation}
    \Phi(x) = \sum_{k=0}^{K}\sum_{q=1}^{Q}T_{k}(\phi(x))[q] \cdot \Omega[q, k],
\label{eq:CKA}
\end{equation}
where the tanh function $\phi(\cdot)$ maps inputs to $\left[-1,1\right]$, required by Chebyshev polynomials, which is essential for preserving their orthogonality and stability during learning.
This polynomial expansion is applied independently to each feature dimension of 
$\phi(x)$, allowing the model to adaptively weight each basis function per dimension. 
Consequently, the classifier can learn rich, dimension-specific nonlinear transformations and approximate complex mappings, thereby enhancing its flexibility to fit diverse data distributions.
As DMIN consists of two branches, we compute classification predictions for the teacher and student branches separately as follows:
\begin{equation}
\begin{cases}
    \widetilde{Y}_{i,HR}^{tea} = [\Phi_{1}(E_{i,HR,1}^{tea}),....,\Phi_{C}(E_{i,HR,C}^{tea})], \\
    \widetilde{Y}_{i,HR}^{stu} = [\Phi_{1}(E_{i,HR,1}^{stu}),....,\Phi_{C}(E_{i,HR,C}^{stu})].
\label{eq:prediction}
\end{cases}
\end{equation}
Here, $\Phi_c(\cdot)$ denotes the output of the CKA classifier corresponding to the $c$-th class.

\subsubsection{Training Objectives}
The training objectives of DMIN are threefold:
1) To ensure that the teacher branch can correctly classify $X_{i,HR}$; 
2) To enforce consistency between the classification results of the student branch (with partial instances) and the teacher branch (with all instances); 
3) To control the proportion of instances selected by the teacher branch. 

Specifically, we use the cross-entropy loss $L_{cls}^{tea}$ to supervise the classification performance of the teacher branch:
\begin{equation}
    L_{cls}^{tea} = CE(\widetilde{Y}_{i,HR}^{tea}, Y_i),
\end{equation}
where $CE(\cdot)$ denotes the cross-entropy loss, and $Y_i$ is the slide-level label of $X_i$.
We also adopt the clustering loss $L_{clu}^{tea}$ from CLAM~\cite{lu2021data} to further structure the feature space in the teacher branch.

Next, we employ knowledge distillation to align the bag-level representation $E_{i,HR}^{stu}$ and classification logits $\widetilde{Y}_{i,HR}^{stu}$ in the student branch with those from the teacher branch:
\begin{equation}
\begin{cases}
    L_{dis,1}^{stu} = L_{2}(E_{i,HR}^{stu}, E_{i,HR}^{tea}), \\
    L_{dis,2}^{stu} = L_{KL}(\widetilde{Y}_{i,HR}^{stu}, \widetilde{Y}_{i,HR}^{tea}).
\end{cases}
\end{equation}
Here, $L_2(\cdot)$ and $L_{KL}(\cdot)$ represent the $\ell_2$ loss and Kullback–Leibler (KL) divergence, respectively.

Finally, we regularize the proportion of selected relevant instances $\widetilde{r}_{i, HR}$ to match a predefined retention ratio $r$:
\begin{equation}
    L_{rate}^{stu} = L_{2}(\widetilde{r}_{i, HR}, r).
\end{equation}
An instance $j$ is considered relevant if $\sum_{c=1}^{C} \mathbb{I}({M}_{i,HR,c}^{j} \ne 0) > 0$, where $\mathbb{I}(\cdot)$ is the indicator function.

The overall loss function of DMIN in the SD training step is defined as:
\begin{align}
    L_{DMIN}^{SD} = & \alpha_{1}L_{cls}^{tea} + \alpha_{2}L_{clu}^{tea} + \alpha_{3}L_{dis,1}^{stu}+ \nonumber\\& \alpha_{4}L_{dis,2}^{stu} + \alpha_{5}L_{rate}^{stu}.
\label{eq:DMIN-loss}
\end{align}
We do not perform hyperparameter tuning for the loss weights. Following CLAM~\cite{lu2021data}, we empirically set $\alpha_1 = 0.7$ and $\alpha_2 = 0.3$. 
The remaining coefficients $\alpha_3 = 0.5$, $\alpha_4 = 0.5$, and $\alpha_5 = 2.0$ are adopted from DynamicViT~\cite{rao2021dynamicvit}.

\subsection{Asymmetric Distillation Training Step}\label{sec:AD}
Although DMIN effectively identifies irrelevant regions in high-resolution WSIs, it does not speed up inference.
This is because it still requires the feature extractor to process all patches to determine which instances should be discarded.
In practice, this exhaustive patch-wise feature extraction constitutes the primary bottleneck in WSI-level inference.
To overcome this, we introduce an AD training step, as illustrated in Figure~\ref{fig:framework}(a), where DMIN and DB-LIPN are trained in a mutually guided manner:
DMIN distills instance-level relevance into DB-LIPN to enable fast region filtering at low resolution, while DB-LIPN guides DMIN's fine-tuning to adapt to the distribution of retained high-resolution patches.

\subsubsection{Cross-Resolution Distillation of DB-LIPN}
Specifically, the $N_i$ 16$\times$16 patches extracted from $X_{i,LR}$ are fed into DB-LIPN, producing two prediction matrices $P_{i,LR,c}^{B1}$ and $P_{i,LR,c}^{B2}$ for each class.
The dual-branch design generates distinct yet synergistic predictions, which are combined to improve robustness, as the branches are independently initialized.
Both branches are then trained under the supervision of ${A}_{i,HR}$, learning to identify irrelevant regions within low-resolution WSIs.
To this end, we introduce two complementary training strategies, detailed below:
\begin{itemize}
    \item \textbf{Hard Mode}. Since low-resolution patches inherently contain limited information, it is challenging for DB-LIPN to replicate DMIN’s ability to assign precise contribution scores to individual instances. 
    Instead, we encourage DB-LIPN to make a simpler binary decision—determining whether each patch contributes to the bag-level classification.
    Accordingly, each element in the prediction matrices $P_{i,LR}^{B1}$ and $P_{i,LR}^{B2}$ is binarized as follows:
    \begin{equation}
    \begin{cases}
        {M}_{i,LR}^{B1,j} = B(P_{i,LR}^{B1,j},\gamma)-D(P_{i,LR}^{B1,j})+P_{i,LR}^{B1,j}, \\
        {M}_{i,LR}^{B2,j} = B(P_{i,LR}^{B2,j},\gamma)-D(P_{i,LR}^{B2,j})+P_{i,LR}^{B2,j}.
    \end{cases}
    \end{equation}
    We then enforce consistency between these binarized predictions and $M_{i,HR}$, enabling the former to effectively indicate instance-level relevance in low-resolution inputs:
    \begin{equation}
        L_{dis,3}=\frac{L_1({M}_{i,LR}^{B1},M_{i,HR})+L_1({M}_{i,LR}^{B2},M_{i,HR})}{2},
    \end{equation}
    where $L1(\cdot)$ denotes the $\ell_1$ loss function.
    \item \textbf{Soft Mode}. Although binary supervision simplifies training, it may cause loss of useful fine-grained relevance cues present in low-resolution patches.
    To better preserve such signals, we propose a soft alternative that directly aligns the prediction matrices $P_{i,LR}^{B1}$ and $P_{i,LR}^{B2}$ with the attention scores $A_{i,HR}$ from DMIN:
    \begin{equation}
        L_{dis,3}=\frac{L_1({P}_{i,LR}^{B1},A_{i,HR})+L_1({P}_{i,LR}^{B2},A_{i,HR})}{2}.
    \end{equation}
\end{itemize}
In practice, to balance the two distillation modes, we select the soft mode with probability $p$ and the hard mode with probability $1-p$ during training. Ablation experiments are conducted to study the effect of $p$ as a hyperparameter. 
For more details, please refer to Section~\ref{sec:Comprehensive Analysis}.

Similar to the training objectives used in the SD step for DMIN, the proportion of predicted relevant patches, denoted as $\widetilde{r}_{i, LR}^{B1}$ and $\widetilde{r}_{i, LR}^{B2}$ for the two branches of DB-LIPN, is also regularized to be close to the predefined retention ratio $r$.
The overall hybrid loss for DB-LIPN in the AD step is defined as:
\begin{equation}
    L_{LIPN}^{AD} = \beta_{1}L_{dis,3} + \beta_{2}\frac{L_{2}(\widetilde{r}_{i, LR}^{B1}, r) + L_{2}(\widetilde{r}_{i, LR}^{B2}, r)}{2}.
\label{eq:LIPN}
\end{equation}

\subsubsection{Cross-Branch Exponential Moving Average}
As shown in Equation~\ref{eq:LIPN}, each branch of DB-LIPN is trained independently with its own relevance ratio constraint, ensuring a fixed proportion of instance retention in each branch.
However, there is no guarantee that both branches will identify the same instances as relevant.
Consequently, when combining the predictions from both branches, the overlap of retained instances may be limited in some cases, leading to a lower overall retention rate.
This issue is particularly concerning for datasets such as Camelyon16, where tumor regions constitute only a small portion of the WSI.
A reduced retention rate after fusion raises the risk of discarding rare yet crucial positive regions, potentially degrading classification performance.
To mitigate this problem, we propose a novel CBEMA strategy.
Specifically, at the end of each training epoch, the parameters of each DB-LIPN branch are additionally updated by blending their own weights with those of the other branch via an exponential moving average, complementing the standard training updates.
Parameters are updated simultaneously via a joint assignment, as shown:
\begin{equation}
    \theta_{1}, \theta_2 = (1 - \lambda)\theta_1+\lambda\theta_2, (1 - \lambda)\theta_2+\lambda\theta_1.
\end{equation}
Here, $\theta_{1}$ and $\theta_2$ denote the learnable parameters of the two branches, and $\lambda$ is a hyperparameter controlling the mixing ratio of the updates.

Compared with other alternative designs, the combination of the dual-branch design and CBEMA offers clear advantages:
\begin{itemize}
    \item \textbf{Over not using CBEMA}. CBEMA enhances consistency between branches by aligning their parameters, leading to a higher overlap in retained instances and reducing the risk of missing critical regions. This is especially beneficial for datasets with sparse positive areas, such as Camelyon16~\cite{bejnordi2017diagnostic}, where overlooking a few relevant patches can significantly degrade performance.
    \item \textbf{Over using a single branch}. The dual-branch design enables the model to capture diverse and complementary discriminative patterns, improving robustness and boosting classification accuracy.
    \item \textbf{Over fusion at the feature map or prediction level}. Parameter-level fusion mitigates intrinsic differences between models more effectively, facilitates knowledge sharing, and avoids the instability commonly introduced by superficial fusion strategies.
\end{itemize}

\subsubsection{Adaptation of DMIN for Efficient Inference}
As illustrated in Figure~\ref{fig:framework}(b), the proposed efficient inference pipeline comprises three steps:
1) Extract all patches from $X_{i,LR}$, generating a total of $N_i$ patches.
2) Input these patches into DB-LIPN to identify regions relevant to classification, yielding  $P_{i,LR}^{B1}$ and $P_{i,LR}^{B2}$;
3) Selectively extract $\widetilde{r}_{i, LR}N_i$ relevant patches from $X_{i,HR}$ based on the merged mask $M_{i,LR}=B(\frac{P_{i,LR}^{B1}+P_{i,LR}^{B2}}{2},\gamma)$, and feed them into the feature extractor and student branch of DMIN for final classification.

To adapt the student branch to the partial inputs selected by DB-LIPN, we fine-tune it using only the selected high-resolution patches:
\begin{equation}
    L_{DMIN}^{AD} = CE(\tilde{Y}_{i,HE}^{stu}, Y_i).
\end{equation}
Throughout the AD training step, DB-LIPN and DMIN are alternately optimized in each iteration via $L_{LIPN}^{AD}$ and $L_{DMIN}^{AD}$, enabling both modules to gradually co-adapt to the evolving instance selection strategy.

\begin{figure}[!h]
    \centering
    \includegraphics[width=0.5\textwidth]{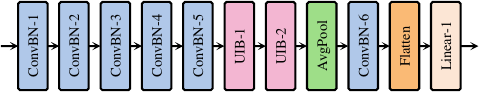}
    \caption{The architecture of a single DB-LIPN branch. ``ConvBN" denotes a sequential module consisting of a convolutional layer, a batch normalization layer, and a ReLU activation. ``UIB" refers to the Universal Inverted Bottleneck proposed in MobileNetV4.}
    \label{fig:architecture_LIPN}
\end{figure} 

\begin{table}[!h]
\centering
\begin{tabular}{c|cc}
\toprule
Layer & Parameter & Output Size \\
\midrule
ConvBN-1 & k=$3$, s=$2$, in=$3$, out=$16$ & $1\times16\times8\times8$ \\
ConvBN-2 & k=$3$, s=$2$, in=$16$, out=$16$ & $1\times16\times4\times4$ \\
ConvBN-3 & k=$1$, s=$1$, in=$16$, out=$16$ & $1\times16\times4\times4$ \\
ConvBN-4 & k=$3$, s=$2$, in=$16$, out=$48$ & $1\times48\times2\times2$ \\
ConvBN-5 & k=$1$, s=$1$, in=$48$, out=$24$ & $1\times24\times2\times2$ \\
UIB-1 & \makecell[c]{sdk=$5$, mdk=$5$, e=$2$,\\s=$2$, in=$24$, out=$48$} & $1\times48\times1\times1$ \\
UIB-2 & \makecell[c]{sdk=$3$, mdk=$3$, e=$2$,\\s=$2$, in=$48$, out=$64$} & $1\times64\times1\times1$ \\
\hline
AvgPool & null & $1\times64\times1\times1$ \\
ConvBN-6 & k=$1$, s=$1$, in=$64$, out=$64$ & $1\times64\times1\times1$ \\
Flatten & null & $1\times64$ \\
Linear-1 & in=$64$, out=$2$ & $1\times2$ \\
\bottomrule
\end{tabular}
\caption{
Detailed layer parameters and output feature dimensions of a single DB-LIPN branch.
``k", ``s", ``in", and ``out" denote the kernel size, stride, number of input channels, and number of output channels, respectively.
``sdk", ``mdk", and ``e" represent the start depthwise kernel size, middle depthwise kernel size, and the expansion ratio, respectively.
The input image has a shape of $1 \times 3 \times 16 \times 16$.
}
\label{tab:parameter_LIPN}
\end{table}

\subsection{Network Architectures}
In our implementation, we adopt a widely used variant of ResNet-50, pre-trained on ImageNet, as the feature extractor, following CLAM~\cite{lu2021data}.
The architecture of the proposed DMIN is illustrated in Figure~\ref{fig:framework}(c).
For each branch of the DB-LIPN module, we employ a lightweight variant of MobileNetV4~\cite{qin2024mobilenetv4}, with its detailed architecture and corresponding parameters provided in Figure~\ref{fig:architecture_LIPN} and Table~\ref{tab:parameter_LIPN}, respectively.

\section{Experiments}\label{sec:experiments}

\subsection{Experimental Settings}
\subsubsection{Datasets and Tasks}
We conducted cancer diagnosis and subtyping experiments on four public pathological WSI datasets spanning three organ sites, aiming to comprehensively evaluate the performance of the proposed method. Specifically:
1) Breast cancer lymph node metastasis detection on Camelyon16~\cite{bejnordi2017diagnostic}, involving normal and tumor WSIs.
2) Non-small cell lung cancer subtyping on TCGA-NSCLC, including lung squamous cell carcinoma (LUSC) and lung adenocarcinoma (LUAD).
3) Breast cancer subtyping on TCGA-BRCA, consisting mainly of invasive ductal carcinoma (IDC) and invasive lobular carcinoma (ILC) WSIs; other WSIs with small sample sizes were excluded following DPBAug~\cite{dong2025disentangled}.
4) Renal cell carcinoma subtyping on TCGA-RCC, covering: renal clear cell carcinoma (KIRC), renal papillary cell carcinoma (KIRP), and chromophobe cell carcinoma (KICH).
Among these tasks, the first three are binary classification problems, while the last one is a three-class classification task.

\begin{table}[htbp]
\caption{
Statistical details of the four publicly available datasets used in this paper.
}
\label{tab:dataset_statistics}
\setlength{\tabcolsep}{2.1pt}
\centering
\begin{tabular}{lccccc}
\toprule
\multirow{2}{*}{Dataset} & \multicolumn{4}{c}{Case} & \multirow{2}{*}{Total WSI}  \\
\cmidrule(r){2-5}
& Train & Validation & Test & Total & \\
\midrule
Camelyon16 & 224 & 26 & 129 & 399 & 399 \\
TCGA-NSCLC & 759 & 93 & 93 & 945 & 1042 \\
TCGA-BRCA & 752 & 93 & 93 & 938 & 997 \\
TCGA-RCC & 708 & 86 & 86 & 880 & 922 \\
\bottomrule
\end{tabular}
\end{table}

\begin{sidewaystable*}[htbp]
\caption{
Comparison of AHDMIL with state-of-the-art MIL methods on the Camelyon16, TCGA-NSCLC, TCGA-BRCA, and TCGA-RCC datasets.
10-fold test AUC, accuracy, F1, and BS are reported as $\mathrm{mean}_{\mathrm{std}}$, with statistical significance against AHDMIL assessed via paired t-tests; results with \textless 0.05 are marked with \ding{51}.
The \textcolor{DeepRed}{\textbf{best}} and \textcolor{DeepBlue}{\textbf{second-best}} results are highlighted.
}
\label{tab:comparative_results}
\setlength{\tabcolsep}{4.4pt}
\centering
\begin{tabular}{lcccccccccccccccc}
\toprule
\multirow{2}{*}{Method} & \multicolumn{8}{c}{Camelyon16} & \multicolumn{8}{c}{TCGA-NSCLC} \\
\cmidrule(r){2-9} \cmidrule(r){10-17}
~ & \multicolumn{2}{c}{AUC $\uparrow$} & \multicolumn{2}{c}{Acc $\uparrow$} & \multicolumn{2}{c}{F1 $\uparrow$} & \multicolumn{2}{c}{BS $\downarrow$} & \multicolumn{2}{c}{AUC $\uparrow$} & \multicolumn{2}{c}{Acc $\uparrow$} & \multicolumn{2}{c}{F1 $\uparrow$} & \multicolumn{2}{c}{BS $\downarrow$} \\
\midrule
Max-Pooling & ${83.26}_{1.54}$ & \ding{51} & ${82.41}_{0.73}$ & \ding{51} & ${71.92}_{1.50}$ & \ding{51} & ${0.14}_{0.00}$ & \ding{51} & ${94.66}_{2.33}$ & \ding{55} & ${86.40}_{3.73}$ & \ding{55} & ${86.21}_{3.87}$ & \ding{51} & ${0.10}_{0.02}$ & \ding{51} \\
Mean-Pooling & ${61.80}_{2.15}$ & \ding{51} & ${70.54}_{1.41}$ & \ding{51} & ${46.12}_{6.25}$ & \ding{51} & ${0.22}_{0.00}$ & \ding{51} & ${92.82}_{3.54}$ & \ding{51} & ${84.93}_{4.78}$ & \ding{51} & ${84.52}_{5.08}$ & \ding{51} & ${0.11}_{0.03}$ & \ding{51} \\
ABMIL~\cite{ilse2018attention} & ${84.88}_{3.38}$ & \ding{51} & ${82.79}_{2.68}$ & \ding{51} & ${75.04}_{4.15}$ & \ding{51} & ${0.14}_{0.02}$ & \ding{51} & ${94.92}_{2.29}$ & \ding{55} & ${88.03}_{3.65}$ & \ding{55} & ${88.36}_{3.58}$ & \ding{55} & ${0.09}_{0.03}$ & \ding{55} \\
CLAM-SB~\cite{lu2021data} & ${83.49}_{4.46}$ & \ding{51} & ${79.61}_{4.40}$ & \ding{51} & ${72.01}_{6.06}$ & \ding{51} & ${0.16}_{0.03}$ & \ding{51} & ${95.05}_{2.72}$ & \ding{55} & ${88.74}_{3.39}$ & \ding{55} & ${88.64}_{3.50}$ & \ding{55} & ${0.09}_{0.02}$ & \ding{55}  \\
CLAM-MB~\cite{lu2021data} & ${87.51}_{3.23}$ & \ding{51} & ${82.56}_{3.11}$ & \ding{51} & ${75.91}_{4.61}$ & \ding{51} & ${0.14}_{0.03}$ & \ding{51}  & ${95.59}_{2.16}$ & \ding{55} & ${88.01}_{3.38}$ & \ding{55} & ${88.03}_{3.43}$ & \ding{55} & ${0.09}_{0.02}$ & \ding{55}  \\
DSMIL~\cite{li2021dual} & ${75.94}_{10.81}$ & \ding{51} & ${75.35}_{6.12}$ & \ding{51} & ${51.82}_{19.26}$ & \ding{51} & ${0.20}_{0.02}$ & \ding{51} & ${92.11}_{2.97}$ & \ding{51} & ${83.67}_{3.80}$ & \ding{51} & ${83.85}_{3.73}$ & \ding{51} & ${0.13}_{0.02}$ & \ding{51}  \\
TransMIL~\cite{shao2021transmil} & ${82.26}_{5.67}$ & \ding{51} & ${81.01}_{6.85}$ & \ding{51} & ${73.58}_{8.21}$ & \ding{51} & ${0.17}_{0.06}$ & \ding{51} & ${94.57}_{2.03}$ & \ding{51} & ${88.21}_{3.04}$ & \ding{51} & ${87.97}_{3.40}$ & \ding{55} & ${0.11}_{0.02}$ & \ding{51} \\
DTFD-AFS~\cite{zhang2022dtfd}  & ${87.40}_{3.17}$ & \ding{51} & ${85.12}_{2.42}$ & \ding{51} & ${80.08}_{2.56}$ & \ding{51} & ${0.13}_{0.01}$ & \ding{51} & ${95.59}_{2.08}$ & \ding{55} & ${88.76}_{3.89}$ & \ding{55} & ${88.83}_{3.73}$ & \ding{55} & ${0.09}_{0.02}$ & \ding{55}  \\
DTFD-MAS~\cite{zhang2022dtfd}  & $\textcolor{DeepBlue}{\textbf{87.75}}_{2.07}$ & \ding{51} & $\textcolor{DeepBlue}{\textbf{85.43}}_{2.03}$ & \ding{51} & $\textcolor{DeepBlue}{\textbf{80.47}}_{2.32}$ & \ding{51} & $\textcolor{DeepBlue}{\textbf{0.12}}_{0.02}$ & \ding{51} & ${95.02}_{2.32}$ & \ding{55} & ${89.02}_{3.78}$ & \ding{55} & ${88.98}_{3.98}$ & \ding{55} & ${0.09}_{0.03}$ & \ding{55}  \\
S4MIL~\cite{fillioux2023structured} & ${86.40}_{1.99}$ & \ding{51} & ${80.39}_{2.79}$ & \ding{51} & ${72.32}_{3.07}$ & \ding{51} & ${0.16}_{0.02}$ & \ding{51} & $\textcolor{DeepBlue}{\textbf{96.19}}_{1.89}$ & \ding{55} & $\textcolor{DeepBlue}{\textbf{89.69}}_{2.86}$ & \ding{55} & ${89.43}_{3.13}$ & \ding{55} & $\textcolor{DeepBlue}{\textbf{0.08}}_{0.02}$ & \ding{55}  \\
MambaMIL~\cite{yang2024mambamil} & ${87.06}_{6.19}$ & \ding{51} & ${83.26}_{2.93}$ & \ding{51} & ${73.98}_{5.18}$ & \ding{51} & ${0.15}_{0.03}$ & \ding{51} & ${95.37}_{1.70}$ & \ding{55} & ${89.62}_{3.13}$ & \ding{55} & $\textcolor{DeepBlue}{\textbf{89.56}}_{3.33}$ & \ding{55} & ${0.09}_{0.02}$ & \ding{55}  \\
\rowcolor[RGB]{255,242,204} AHDMIL & $\textcolor{DeepRed}{\textbf{91.96}}_{1.29}$ & - & $\textcolor{DeepRed}{\textbf{89.92}}_{1.51}$ & - & $\textcolor{DeepRed}{\textbf{85.23}}_{2.39}$ & - & $\textcolor{DeepRed}{\textbf{0.09}}_{0.01}$ & - & $\textcolor{DeepRed}{\textbf{96.64}}_{1.93}$ & - & $\textcolor{DeepRed}{\textbf{90.72}}_{2.99}$ & - & $\textcolor{DeepRed}{\textbf{90.70}}_{3.03}$ & - & $\textcolor{DeepRed}{\textbf{0.07}}_{0.02}$ & -  \\
\midrule
\midrule
\multirow{2}{*}{Method} & \multicolumn{8}{c}{TCGA-BRCA} & \multicolumn{8}{c}{TCGA-RCC} \\
\cmidrule(r){2-9} \cmidrule(r){10-17}
~ & \multicolumn{2}{c}{AUC $\uparrow$} & \multicolumn{2}{c}{Acc $\uparrow$} & \multicolumn{2}{c}{F1 $\uparrow$} & \multicolumn{2}{c}{BS $\downarrow$} & \multicolumn{2}{c}{AUC $\uparrow$} & \multicolumn{2}{c}{Acc $\uparrow$} & \multicolumn{2}{c}{F1 $\uparrow$} & \multicolumn{2}{c}{BS $\downarrow$} \\
\midrule
Max-Pooling & ${88.03}_{7.76}$ & \ding{55} & ${86.05}_{3.88}$ & \ding{55} & ${63.18}_{11.62}$ & \ding{51} & ${0.10}_{0.03}$ & \ding{55} & ${98.39}_{1.01}$ & \ding{55} & ${92.17}_{3.65}$ & \ding{55} & ${90.37}_{4.05}$ & \ding{55} & ${0.12}_{0.05}$ & \ding{55} \\
Mean-Pooling & ${88.23}_{5.67}$ & \ding{55} & ${86.74}_{2.44}$ & \ding{55} & ${62.32}_{12.63}$ & \ding{51} & ${0.10}_{0.02}$ & \ding{55}  & ${97.91}_{1.87}$ & \ding{55} & ${90.69}_{5.40}$ & \ding{55} & ${88.41}_{6.15}$ & \ding{55} & ${0.13}_{0.08}$ & \ding{55}  \\
ABMIL~\cite{ilse2018attention} & ${87.70}_{6.15}$ & \ding{55} & ${87.68}_{3.51}$ & \ding{55} & ${67.41}_{10.29}$ & \ding{55} & ${0.10}_{0.03}$ & \ding{55}  & ${98.62}_{1.33}$ & \ding{55} & ${92.88}_{3.62}$ & \ding{55} & ${91.56}_{4.12}$ & \ding{55} & ${0.11}_{0.06}$ & \ding{55}  \\
CLAM-SB~\cite{lu2021data} & ${88.25}_{6.12}$ & \ding{55} & ${87.58}_{4.92}$ & \ding{55} & ${68.44}_{12.29}$ & \ding{55} & ${0.10}_{0.04}$ & \ding{55}  & ${98.55}_{1.32}$ & \ding{55} & ${91.62}_{4.18}$ & \ding{55} & ${89.53}_{5.01}$ & \ding{55} & ${0.12}_{0.07}$ & \ding{55}  \\
CLAM-MB~\cite{lu2021data} & ${90.22}_{5.18}$ & \ding{55} & $\textcolor{DeepBlue}{\textbf{88.27}}_{3.52}$ & \ding{55} & $\textcolor{DeepBlue}{\textbf{70.29}}_{8.89}$ & \ding{55} & ${0.09}_{0.02}$ & \ding{55}  & ${98.76}_{1.14}$ & \ding{55} & ${92.20}_{4.65}$ & \ding{55} & ${90.54}_{6.03}$ & \ding{55} & ${0.11}_{0.06}$ & \ding{55}  \\
DSMIL~\cite{li2021dual} & ${83.33}_{7.48}$ & \ding{51} & ${82.59}_{3.66}$ & \ding{51} & ${55.34}_{10.29}$ & \ding{51} & ${0.14}_{0.02}$ & \ding{51} & ${97.84}_{1.38}$ & \ding{51} & ${89.81}_{4.82}$ & \ding{51} & ${87.62}_{5.62}$ & \ding{51} & ${0.17}_{0.05}$ & \ding{51} \\
TransMIL~\cite{shao2021transmil} & ${88.33}_{5.73}$ & \ding{55} & ${87.55}_{3.78}$ & \ding{55} & ${67.60}_{9.07}$ & \ding{55} & ${0.11}_{0.04}$ & \ding{55}  & ${98.21}_{0.93}$ & \ding{51} & ${92.31}_{3.52}$ & \ding{51} & ${90.01}_{4.46}$ & \ding{55} & ${0.13}_{0.05}$ & \ding{55} \\
DTFD-AFS~\cite{zhang2022dtfd}  & ${87.24}_{7.38}$ & \ding{55} & ${86.83}_{3.98}$ & \ding{55} & ${65.80}_{11.76}$ & \ding{55} & ${0.11}_{0.03}$ & \ding{55}  & ${98.49}_{1.24}$ & \ding{55} & ${92.30}_{4.52}$ & \ding{55} & ${90.97}_{4.55}$ & \ding{55} & ${0.12}_{0.07}$ & \ding{55}  \\
DTFD-MAS~\cite{zhang2022dtfd} & ${87.80}_{9.65}$ & \ding{55} & ${87.48}_{4.13}$ & \ding{55} & ${66.95}_{12.21}$ & \ding{55} & ${0.10}_{0.03}$ & \ding{55}  & ${98.22}_{1.59}$ & \ding{55} & ${92.41}_{4.82}$ & \ding{55} & ${91.33}_{5.68}$ & \ding{55} & ${0.12}_{0.07}$ & \ding{55}  \\
S4MIL~\cite{fillioux2023structured} & $\textcolor{DeepBlue}{\textbf{90.40}}_{5.73}$ & \ding{55} & ${88.17}_{3.88}$ &  \ding{55} & ${65.07}_{13.50}$ &  \ding{55} & $\textcolor{DeepBlue}{\textbf{0.09}}_{0.03}$ & \ding{55} & $\textbf{\textcolor{DeepBlue}{99.02}}_{0.99}$ & \ding{55} & ${92.98}_{4.26}$ & \ding{55} & ${91.44}_{5.14}$ & \ding{55} & $\textcolor{DeepRed}{\textbf{0.10}}_{0.06}$ & \ding{55}  \\
MambaMIL~\cite{yang2024mambamil} & ${89.69}_{5.91}$ & \ding{55} & ${87.78}_{4.27}$ & \ding{55} & ${69.25}_{9.64}$ & \ding{55} & ${0.10}_{0.03}$ & \ding{55}  & ${98.45}_{1.14}$ & \ding{55} & $\textbf{\textcolor{DeepBlue}{93.32}}_{2.66}$ & \ding{55} & $\textbf{\textcolor{DeepBlue}{91.92}}_{3.23}$ & \ding{55} & ${0.11}_{0.05}$ & \ding{55}  \\
\rowcolor[RGB]{255,242,204} AHDMIL & $\textcolor{DeepRed}{\textbf{90.93}}_{5.21}$ & - & $\textcolor{DeepRed}{\textbf{88.87}}_{2.35}$ & - & $\textcolor{DeepRed}{\textbf{71.89}}_{5.98}$ & - & $\textcolor{DeepRed}{\textbf{0.09}}_{0.02}$ & - & $\textcolor{DeepRed}{\textbf{99.02}}_{0.61}$ & - & $\textcolor{DeepRed}{\textbf{93.78}}_{1.99}$ & - & $\textcolor{DeepRed}{\textbf{92.61}}_{2.26}$ & -  & $\textcolor{DeepBlue}{\textbf{0.11}}_{0.03}$ & -  \\
\bottomrule
\end{tabular}
\end{sidewaystable*}

\subsubsection{Paired-Resolution Patch Tiling}
\label{sec:PRPT}
All WSIs were pre-processed using tools developed in CLAM~\cite{lu2021data}.
As previously mentioned, 20$\times$ WSIs are used as high-resolution inputs and 1.25$\times$ WSIs as low-resolution inputs in our experiments.
To ensure strict spatial correspondence between patches cropped  from the two resolutions, we first perform tissue segmentation on the 20$\times$ WSIs and divide them into non-overlapping 256$\times$256 patches.
During this process, we record the spatial coordinates of each patch and map them to their corresponding positions on the 1.25$\times$ WSIs.
Using the mapped coordinates, we then extract 16$\times$16 patches from the 1.25$\times$ WSIs, ensuring precise spatial alignment between resolutions.
WSIs with insufficient tissue area are discarded during pre-processing.
It is worth noting that WSIs in the TCGA project occasionally suffer from \textit{resolution omissions}—that is, some slides may lack 20$\times$ or 1.25$\times$ versions but still contain 40$\times$ and 5$\times$ scans.
To address this, we follow a strategy similar to TransMIL~\cite{shao2021transmil}: 512$\times$512 or 32$\times$32 patches are extracted from the corresponding positions on 40$\times$ or 5$\times$ WSIs, respectively, and downsampled by a factor of 2 using bicubic interpolation.
This approach helps maintain sufficient data availability for training and evaluation.

\subsubsection{Cross-validation Protocol} 
All experiments followed a 10-fold Monte Carlo cross-validation protocol.
Since TCGA datasets often contain multiple WSIs per patient case, to prevent information leakage, we performed dataset splitting at the case level rather than the slide level.
Specifically:
1) For the Camelyon16 dataset, the official training set was further split into training and validation subsets in a 9:1 ratio by case, while the official test set was  used consistently across all folds. 
2) For the TCGA-NSCLC, TCGA-BRCA, and TCGA-RCC datasets, cases were split into training, validation, and test sets using an 8:1:1 ratio within each fold.
Details regarding numbers of WSIs and cases, and dataset splits for each dataset are summarized in Table~\ref{tab:dataset_statistics}.

\subsubsection{Evaluation Metrics}
To comprehensively assess classification performance, we employ four widely used metrics:
\begin{itemize}
\item \textbf{AUC} is a common metric for evaluating binary classifiers. 
It measures the probability that the model assigns a higher score to a randomly chosen positive sample than to a randomly chosen negative sample.
For multiclass problems, we adopt Macro-AUC, which calculates the one-vs-rest AUC for each class and averages these scores, providing an overall measure of the model’s ability to rank the true class higher than others across all categories.

\item \textbf{Accuracy (Acc)} measures the overall proportion of correctly predicted samples across all classes, reflecting the model’s general performance. 
Its calculation is identical for both binary and multiclass classification tasks.

\item \textbf{Macro f1-score (F1)} is computed by calculating the harmonic mean of precision and recall for each class independently, then averaging the results with equal weight for all classes.
In binary classification, Macro F1 is equivalent to the standard F1-score of the positive class.

\item \textbf{Brier score (BS)} measures the calibration quality of probabilistic predictions by computing the mean squared error between predicted probabilities and true labels.
For binary classification, the computation is performed on the predicted probability for the positive class and the corresponding ground truth label.
For multiclass classification, it is performed on the full predicted probability vector and the one-hot encoded ground truth across all classes.
Lower BS indicates better calibration.

\end{itemize}
To ensure consistency across both binary and multiclass settings, we report Macro-AUC and Macro-F1 throughout this paper. AUC, Acc, and F1 are expressed in percentage format.

\subsubsection{Implementation Details}
For the training of both DMIN and DB-LIPN, we use the Adam~\cite{kingma2014adam} optimizer with a batch size of 1. 
The SD learning rate settings, chebyshev polynomial degree $K$, and predefined instance retention ratio $p$ used in this paper and  are consistent with those in our earlier conference version~\cite{dong2025fast}: the learning rate equals 3e-4 for Camelyon16, TCGA-BRCA, TCGA-RCC and 3e-5 for TCGA-NSCLC; $K$ equals 12 for Camelyon16, TCGA-NSCLC, TCGA-RCC and 16 for TCGA-NSCLC; $r$ equals 0.6 for Camelyon16 and TCGA-RCC, while equals 0.7 for TCGA-NSCLC and TCGA-BRCA;
For the AD step, the learning rate is set to 1e-5 for both the DMIN fine-tuning and DB-LIPN training. 
The reduced feature dimensionality $Q$ is set to 512.
The hyper-parameters $\tau$ and $\gamma$ in Equation~\ref{eq:gumbel-sigmoid}, and Equation~\ref{eq:discrete-mask} are fixed at 0.7 and 0.5, respectively, for all datasets.
The coefficients $\beta_1$ and $\beta_2$ are both set to 1.0.
$\lambda$ are set to 0.2 for the Camelyon16 dataset and 0.0 for the TCGA-NSCLC, TCGA-BRCA and TCGA-RCC datasets.
For further implementation details, please refer to our open-source code at  https://github.com/JiuyangDong/AHDMIL.

\subsection{Comparison with State-of-the-Art Methods}

\begin{figure*}[htbp]
    \centering
    \includegraphics[width=1.0\textwidth]{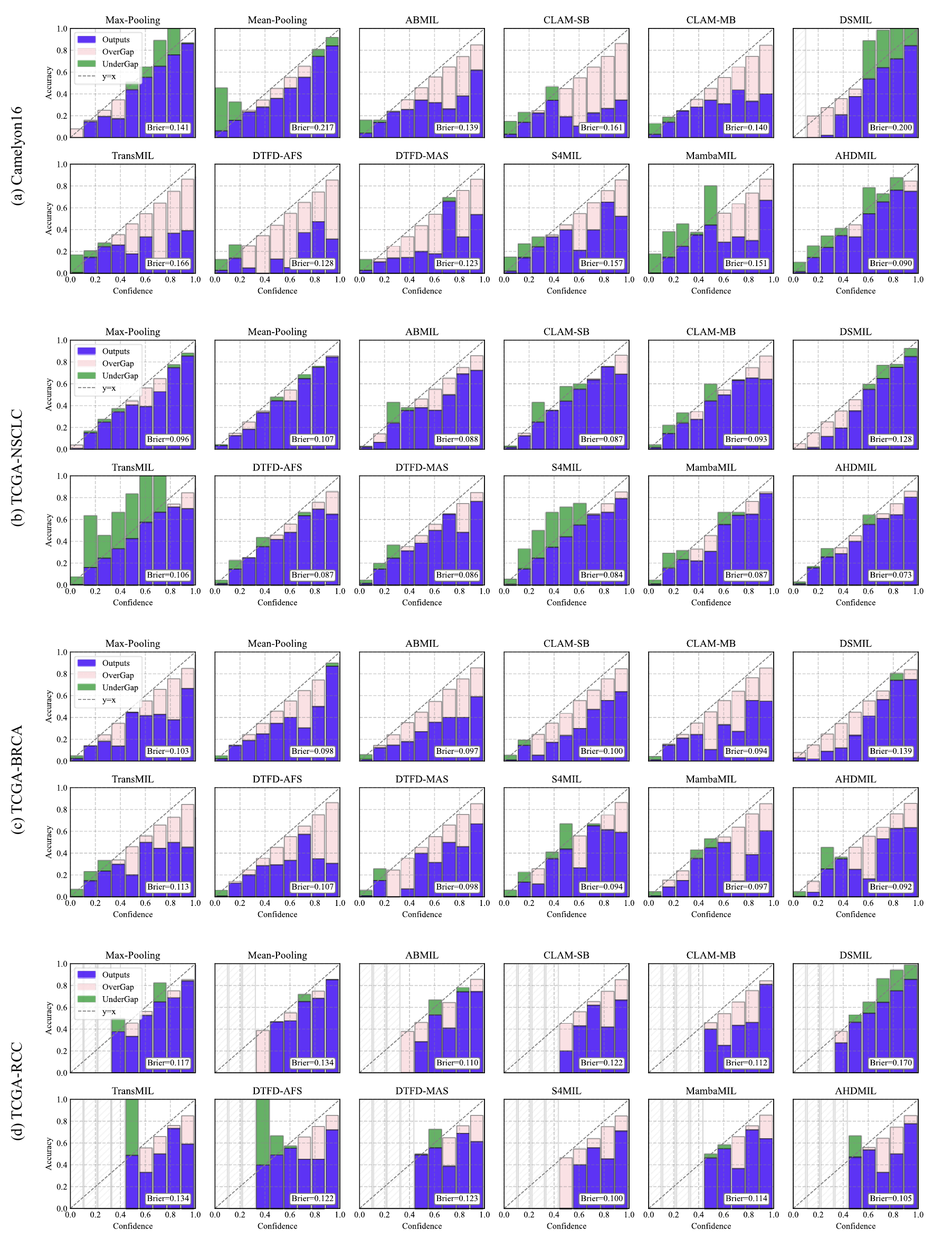}
    \caption{
    Calibration curves of different methods on the Camelyon16, TCGA-NSCLC, TCGA-BRCA, and TCGAs datasets.
    }
    \label{fig:calibration_curve}
\end{figure*}

\subsubsection{Comparative Methods}
We conduct a systematic comparison between our proposed AHDMIL framework and 11 variants from nine state-of-the-art MIL methods, evaluating classification performance, calibration quality, and inference efficiency. The compared methods include Max-Pooling, Mean-Pooling, ABMIL~\cite{ilse2018attention}, CLAM-CB~\cite{lu2021data}, CLAM-MB~\cite{lu2021data}, DSMIL~\cite{li2021dual}, TransMIL~\cite{shao2021transmil}, DTFD-AFS~\cite{zhang2022dtfd}, DTFD-MAS~\cite{zhang2022dtfd}, S4MIL~\cite{fillioux2023structured}, and MambaMIL~\cite{yang2024mambamil}.

For fair comparison, all methods, including AHDMIL, adopt the modified ResNet-50 backbone as the feature extractor following CLAM. Multi-resolution feature concatenation is not used in any method during inference. While the original DSMIL implementation leverages self-supervised features and multi-scale fusion, we disable these components to ensure consistency and isolate the contributions of the MIL algorithms themselves.

\subsubsection{Classification Performance}
Table~\ref{tab:comparative_results} presents the classification performance of various MIL methods across the 10-fold test sets of the Camelyon16, TCGA-NSCLC, TCGA-BRCA, and TCGA-RCC datasets. From the table, we draw the following key observations:
\begin{itemize}
    \item AHDMIL consistently achieves performance that is superior to or on par with the best-performing comparative methods across all evaluation metrics. The improvement is particularly pronounced on the Camelyon16 dataset, where AHDMIL achieves increases of 4.8\%, 5.3\%, and 5.9\% in AUC, accuracy, and F1-score, respectively, along with a decrease of approximately 0.03 in Brier score. On the other three datasets, while the baseline methods already perform strongly (with AUCs over 90\% in the SOTA methods), AHDMIL still shows steady and competitive gains.
    \item In terms of statistical significance, the trend aligns with the overall performance results. On the Camelyon16 dataset, AHDMIL achieves statistically significant improvements over all baseline methods across all metrics. For the other three datasets, although the improvements are generally not statistically significant, AHDMIL consistently attains the highest average scores, indicating stable and favorable performance.
    \item Except for AHDMIL, none of the comparative methods exhibit dominant performance across all datasets. For example, DTFD-AFS outperforms all other methods (except AHDMIL) on the Camelyon16 dataset, likely due to its pseudo-bag based data augmentation mitigating the limited sample size issue. However, it performs less competitively on the TCGA-RCC dataset. Such performance variations among baseline methods indirectly emphasize the stability and robustness of AHDMIL across datasets with different characteristics.
\end{itemize}
Taken together, these findings suggest that AHDMIL not only delivers strong overall classification performance but also demonstrates remarkable robustness across diverse WSI datasets. Unlike other methods that perform well only in specific settings, AHDMIL consistently outperforms all baselines, underscoring its adaptability and practical value in real-world application.

\subsubsection{Calibration Quality}
Beyond classification accuracy, reliable prediction confidence is equally important for practical WSI analysis.
To evaluate this, we employ calibration curves to compare the calibration quality of AHDMIL against other methods, as shown in Figure~\ref{fig:calibration_curve}.
Calibration curves are generated by aggregating all test samples across the 10-fold splits.
For binary classification, the curves use predicted confidence scores of the positive class; for multi-class tasks, the maximum predicted probability (i.e., confidence of the predicted class) is used.
Each curve plots observed frequencies against predicted confidence; the closer it lies to the diagonal ($y = x$), the better the model’s calibration.

As shown in Figure~\ref{fig:calibration_curve}, AHDMIL consistently demonstrates superior and stable calibration performance across all datasets, achieving the lowest or second-lowest Brier scores. 
We further visualize the gaps between predicted confidence and observed frequencies within different confidence intervals for each model: overconfidence-related gaps are marked with pink squares, while underconfidence-related gaps are marked with green squares.
This visualization reveals several noteworthy observations: 
\begin{itemize}
    \item On the Camelyon16 and TCGA-BRCA datasets, many models exhibit prevalent overconfidence, likely due to limited training samples in Camelyon16 and class imbalance in TCGA-BRCA. These factors may cause overfitting, leading to overconfident predictions and notable gaps from the ideal diagonal on calibration curves.
    \item On the TCGA-NSCLC dataset, TransMIL and S4MIL exhibit evident underconfidence, unlike other methods. This suggests that different MIL paradigms may differently affect confidence calibration when handling complex data distributions, reflecting variation in probabilistic estimation across methods.
    \item On the TCGA-RCC dataset, nearly all models’ prediction confidences concentrate in higher intervals, with few low-confidence predictions. Given the relatively balanced class distribution and high classification accuracy (all above 90\%) across methods, this suggests that WSI subtypes in TCGA-RCC are more separable in feature space, allowing models to learn clearer decision boundaries.
\end{itemize}
These observations reveal the nuanced calibration behaviors of different methods across datasets, while highlighting AHDMIL’s robustness and superior calibration consistency, evidenced by its low Brier scores and fewer calibration gaps.

\subsubsection{Inference Time}
We divide the overall pipeline for WSI classification using MIL models into two primary stages: preprocessing and model inference.
In the standard preprocessing stage, each WSI is first segmented to extract foreground regions, then tiled into multiple patches. 
Feature representations for each patch are extracted via a pretrained encoder and stored on disk.
The model inference stage consists of three steps: loading the trained MIL model into memory, retrieving the pre-extracted patch features from disk, and generating predictions by passing the features through the model.
In our proposed AHDMIL framework, the patching process is extended to include additional low-resolution crops of the WSI. 
These low-resolution patches are also utilized during model inference to assist and enhance the model's decision-making process.

To compare the processing efficiency of different methods, we report the average time spent on each step of the pipeline for each method, as shown in Table~\ref{tab:MIl_inference_time}.
Specifically, $T_{Seg}$, $T_{Til}$, $T_{FE}$, $T_{FL}$, $T_{ML}$, and $T_{Infer}$  denote the average time per WSI for foreground segmentation, patch tiling, feature extraction, feature loading, model loading, and model inference, respectively. The total processing time per WSI is denoted as $T_{Total}$.
It is worth noting that patch tiling only requires directly accessing patches from the WSI using coordinate-based indexing, followed by immediate input to the feature extractor.
Finally, we note that the reported model processing times differ from those in our earlier conference version. While the conference version averaged the model loading time across all test samples, here we adopt a stricter evaluation protocol in which the model is reloaded for each WSI. This setting more accurately reflects real-world deployment latency, particularly in cold-start scenarios. All inference time measurements were conducted on a machine equipped with an NVIDIA RTX 3090 GPU and an Intel Core i9-14900K CPU, using a batch size of 1024 for feature extraction.

Table~\ref{tab:MIl_inference_time} clearly shows that $T_{Seg}$ is identical across all methods.
For existing methods, the time required for patch tiling and feature extraction is similarly consistent.
In contrast, our proposed AHDMIL framework achieves varying levels of acceleration in these steps, depending on the number of relevant instances retained within each WSI.
The remaining steps—feature loading, model loading, and model inference—require negligible processing time.
Overall, AHDMIL delivers speedups of 1.2$\times$, 1.3$\times$, 1.3$\times$, and 2.1$\times$ on the Camelyon16, TCGA-NSCLC, TCGA-BRCA, and TCGA-RCC datasets, respectively.
These findings indicate that, in addition to consistently outperforming state-of-the-art methods in classification accuracy, AHDMIL also provides significantly faster inference.

\begin{sidewaystable*}[htbp]
\caption{
Average processing time per WSI (in seconds), calculated from 10-fold test sets of the four datasets.
$T_{Seg}$, $T_{Til}$, $T_{FE}$, $T_{FL}$, $T_{ML}$, and $T_{Infer}$ represent the average times for foreground segmentation, patch tiling, feature extraction, feature loading, model loading, and model inference on each WSI, respectively.
$T_{Total}$ denotes the total average processing time per WSI.
Cells highlighted in blue indicate that the comparative methods have identical $T_{Seg}$, $T_{Til}$, $T_{FE}$, and $T_{FL}$ times.
}
\label{tab:MIl_inference_time}
\setlength{\tabcolsep}{5pt}
\centering
\begin{tabular}{lcccccccccccccc}
\toprule
\multirow{2}{*}{Method} & \multicolumn{7}{c}{Camelyon16} & \multicolumn{7}{c}{TCGA-NSCLC} \\
\cmidrule(r){2-8} \cmidrule(r){9-15}
~ & $T_{Seg}$ & $T_{Til}$ & $T_{FE}$ & $T_{FL}$ & $T_{ML}$ & $T_{Infer}$ & \cellcolor[RGB]{230, 255, 230}$T_{Total}$ & $T_{Seg}$ & $T_{Til}$ & $T_{FE}$ & $T_{FL}$ & $T_{ML}$ & $T_{Infer}$ & \cellcolor[RGB]{230, 255, 230}$T_{Total}$ \\
\midrule
Max-Pooling & \cellcolor[RGB]{204, 229, 255} & \cellcolor[RGB]{204, 229, 255} & \cellcolor[RGB]{204, 229, 255} & \cellcolor[RGB]{204, 229, 255} & $0.002$ & $0.008$ & \cellcolor[RGB]{230, 255, 230}$23.462$ & \cellcolor[RGB]{204, 229, 255} & \cellcolor[RGB]{204, 229, 255} & \cellcolor[RGB]{204, 229, 255} & \cellcolor[RGB]{204, 229, 255} & $0.002$ & $0.008$ & \cellcolor[RGB]{230, 255, 230}$59.696$ \\
Mean-Pooling & \cellcolor[RGB]{204, 229, 255} & \cellcolor[RGB]{204, 229, 255} & \cellcolor[RGB]{204, 229, 255} & \cellcolor[RGB]{204, 229, 255} & $0.002$ & $0.008$ & \cellcolor[RGB]{230, 255, 230}$23.462$ & \cellcolor[RGB]{204, 229, 255} & \cellcolor[RGB]{204, 229, 255} & \cellcolor[RGB]{204, 229, 255} & \cellcolor[RGB]{204, 229, 255} & $0.002$ & $0.008$ & \cellcolor[RGB]{230, 255, 230}$59.696$ \\
ABMIL~\cite{ilse2018attention} & \cellcolor[RGB]{204, 229, 255} & \cellcolor[RGB]{204, 229, 255} & \cellcolor[RGB]{204, 229, 255} & \cellcolor[RGB]{204, 229, 255} & $0.002$ & $0.008$ & \cellcolor[RGB]{230, 255, 230}$23.462$& \cellcolor[RGB]{204, 229, 255} & \cellcolor[RGB]{204, 229, 255} & \cellcolor[RGB]{204, 229, 255} & \cellcolor[RGB]{204, 229, 255} & $0.002$ & $0.008$ & \cellcolor[RGB]{230, 255, 230}$59.696$ \\
CLAM-SB~\cite{lu2021data} & \cellcolor[RGB]{204, 229, 255} & \cellcolor[RGB]{204, 229, 255} & \cellcolor[RGB]{204, 229, 255} & \cellcolor[RGB]{204, 229, 255} & $0.006$ & $0.008$ & \cellcolor[RGB]{230, 255, 230}$23.466$ & \cellcolor[RGB]{204, 229, 255} & \cellcolor[RGB]{204, 229, 255} & \cellcolor[RGB]{204, 229, 255} & \cellcolor[RGB]{204, 229, 255} & $0.006$ & $0.008$ & \cellcolor[RGB]{230, 255, 230}$59.700$ \\
CLAM-MB~\cite{lu2021data} & \cellcolor[RGB]{204, 229, 255} & \cellcolor[RGB]{204, 229, 255} & \cellcolor[RGB]{204, 229, 255} & \cellcolor[RGB]{204, 229, 255} & $0.005$ & $0.009$ & \cellcolor[RGB]{230, 255, 230}$23.466$ & \cellcolor[RGB]{204, 229, 255} & \cellcolor[RGB]{204, 229, 255} & \cellcolor[RGB]{204, 229, 255} & \cellcolor[RGB]{204, 229, 255} & $0.004$ & $0.010$ & \cellcolor[RGB]{230, 255, 230}$59.700$ \\
DSMIL~\cite{li2021dual} & \cellcolor[RGB]{204, 229, 255}$1.584$ & \cellcolor[RGB]{204, 229, 255}$11.861$ & \cellcolor[RGB]{204, 229, 255}$9.995$ & \cellcolor[RGB]{204, 229, 255}$0.012$ & $0.008$ & $0.009$ & \cellcolor[RGB]{230, 255, 230}$23.469$  & \cellcolor[RGB]{204, 229, 255}$2.459$ & \cellcolor[RGB]{204, 229, 255}$47.093$ & \cellcolor[RGB]{204, 229, 255}$10.122$ & \cellcolor[RGB]{204, 229, 255}$0.012$ & $0.008$ & $0.009$ & \cellcolor[RGB]{230, 255, 230}$59.703$ \\
TransMIL~\cite{shao2021transmil} & \cellcolor[RGB]{204, 229, 255} & \cellcolor[RGB]{204, 229, 255} & \cellcolor[RGB]{204, 229, 255} & \cellcolor[RGB]{204, 229, 255} & $0.031$ & $0.010$ & \cellcolor[RGB]{230, 255, 230}$23.493$ & \cellcolor[RGB]{204, 229, 255} & \cellcolor[RGB]{204, 229, 255} & \cellcolor[RGB]{204, 229, 255} & \cellcolor[RGB]{204, 229, 255} & $0.030$ & $0.009$ & \cellcolor[RGB]{230, 255, 230}$59.725$ \\
DTFD-AFS~\cite{zhang2022dtfd} & \cellcolor[RGB]{204, 229, 255} & \cellcolor[RGB]{204, 229, 255} & \cellcolor[RGB]{204, 229, 255} & \cellcolor[RGB]{204, 229, 255} & $0.003$ & $0.008$ & \cellcolor[RGB]{230, 255, 230}$23.463$ & \cellcolor[RGB]{204, 229, 255} & \cellcolor[RGB]{204, 229, 255} & \cellcolor[RGB]{204, 229, 255} & \cellcolor[RGB]{204, 229, 255} & $0.003$ & $0.008$ & \cellcolor[RGB]{230, 255, 230}$59.697$ \\
DTFD-MAS~\cite{zhang2022dtfd} & \cellcolor[RGB]{204, 229, 255} & \cellcolor[RGB]{204, 229, 255} & \cellcolor[RGB]{204, 229, 255} & \cellcolor[RGB]{204, 229, 255} & $0.005$ & $0.010$ & \cellcolor[RGB]{230, 255, 230}$23.467$ & \cellcolor[RGB]{204, 229, 255} & \cellcolor[RGB]{204, 229, 255} & \cellcolor[RGB]{204, 229, 255} & \cellcolor[RGB]{204, 229, 255} & $0.004$ & $0.010$ & \cellcolor[RGB]{230, 255, 230}$59.700$ \\
S4MIL~\cite{fillioux2023structured} & \cellcolor[RGB]{204, 229, 255} & \cellcolor[RGB]{204, 229, 255}& \cellcolor[RGB]{204, 229, 255} & \cellcolor[RGB]{204, 229, 255} & $0.011$ & $0.011$ & \cellcolor[RGB]{230, 255, 230}$23.474$ & \cellcolor[RGB]{204, 229, 255} & \cellcolor[RGB]{204, 229, 255} & \cellcolor[RGB]{204, 229, 255} & \cellcolor[RGB]{204, 229, 255} & $0.007$ & $0.015$ & \cellcolor[RGB]{230, 255, 230}$59.708$ \\
MambaMIL~\cite{yang2024mambamil} & \cellcolor[RGB]{204, 229, 255} & \cellcolor[RGB]{204, 229, 255} & \cellcolor[RGB]{204, 229, 255} & \cellcolor[RGB]{204, 229, 255} & $0.057$ & $0.009$ & \cellcolor[RGB]{230, 255, 230}$23.518$  & \cellcolor[RGB]{204, 229, 255} & \cellcolor[RGB]{204, 229, 255} & \cellcolor[RGB]{204, 229, 255} & \cellcolor[RGB]{204, 229, 255} & $0.056$ & $0.009$ & \cellcolor[RGB]{230, 255, 230}$59.751$ \\
\rowcolor[RGB]{255,242,204}AHDMIL & $1.584$ & $10.566$ & $6.884$ & $0.015$ & $0.015$ & $0.023$ & \cellcolor[RGB]{230, 255, 230}$19.087$ & $2.459$ & $35.501$ & $6.340$ & $0.014$ & $0.015$ & $0.029$ & $44.358$ \\
\midrule
\midrule
\multirow{2}{*}{Method} & \multicolumn{7}{c}{TCGA-BRCA} & \multicolumn{7}{c}{TCGA-RCC} \\
\cmidrule(r){2-8} \cmidrule(r){9-15}
~ & $T_{Seg}$ & $T_{Til}$ & $T_{FE}$ & $T_{FL}$ & $T_{ML}$ & $T_{Infer}$ & \cellcolor[RGB]{230, 255, 230}$T_{Total}$ & $T_{Seg}$ & $T_{Til}$ & $T_{FE}$ & $T_{FL}$ & $T_{ML}$ & $T_{Infer}$ & \cellcolor[RGB]{230, 255, 230}$T_{Total}$ \\
\midrule
Max-Pooling & \cellcolor[RGB]{204, 229, 255} & \cellcolor[RGB]{204, 229, 255} & \cellcolor[RGB]{204, 229, 255} & \cellcolor[RGB]{204, 229, 255} & $0.002$ & $0.007$ & \cellcolor[RGB]{230, 255, 230}$36.478$ & \cellcolor[RGB]{204, 229, 255} & \cellcolor[RGB]{204, 229, 255} & \cellcolor[RGB]{204, 229, 255} & \cellcolor[RGB]{204, 229, 255} & $0.002$ & $0.009$ & \cellcolor[RGB]{230, 255, 230}$55.035$ \\
Mean-Pooling & \cellcolor[RGB]{204, 229, 255} & \cellcolor[RGB]{204, 229, 255} & \cellcolor[RGB]{204, 229, 255} & \cellcolor[RGB]{204, 229, 255} & $0.002$ & $0.007$ & \cellcolor[RGB]{230, 255, 230}$36.478$ & \cellcolor[RGB]{204, 229, 255} & \cellcolor[RGB]{204, 229, 255} & \cellcolor[RGB]{204, 229, 255} & \cellcolor[RGB]{204, 229, 255} & $0.002$ & $0.009$ & \cellcolor[RGB]{230, 255, 230}$55.035$ \\
ABMIL~\cite{ilse2018attention} & \cellcolor[RGB]{204, 229, 255} & \cellcolor[RGB]{204, 229, 255} & \cellcolor[RGB]{204, 229, 255} & \cellcolor[RGB]{204, 229, 255} & $0.002$ & $0.008$ & \cellcolor[RGB]{230, 255, 230}$36.479$ & \cellcolor[RGB]{204, 229, 255} & \cellcolor[RGB]{204, 229, 255} & \cellcolor[RGB]{204, 229, 255} & \cellcolor[RGB]{204, 229, 255} & $0.002$ & $0.009$ & \cellcolor[RGB]{230, 255, 230}$55.035$ \\
CLAM-SB~\cite{lu2021data} & \cellcolor[RGB]{204, 229, 255} & \cellcolor[RGB]{204, 229, 255} & \cellcolor[RGB]{204, 229, 255} & \cellcolor[RGB]{204, 229, 255} & $0.005$ & $0.008$ & \cellcolor[RGB]{230, 255, 230}$36.482$ & \cellcolor[RGB]{204, 229, 255} & \cellcolor[RGB]{204, 229, 255} & \cellcolor[RGB]{204, 229, 255} & \cellcolor[RGB]{204, 229, 255} & $0.006$ & $0.009$ & \cellcolor[RGB]{230, 255, 230}$55.039$ \\
CLAM-MB~\cite{lu2021data} & \cellcolor[RGB]{204, 229, 255} & \cellcolor[RGB]{204, 229, 255} & \cellcolor[RGB]{204, 229, 255} & \cellcolor[RGB]{204, 229, 255} & $0.004$ & $0.009$ & \cellcolor[RGB]{230, 255, 230}$36.482$ & \cellcolor[RGB]{204, 229, 255} & \cellcolor[RGB]{204, 229, 255} & \cellcolor[RGB]{204, 229, 255} & \cellcolor[RGB]{204, 229, 255} & $0.005$ & $0.011$ & \cellcolor[RGB]{230, 255, 230}$55.040$ \\
DSMIL~\cite{li2021dual} & \cellcolor[RGB]{204, 229, 255}$2.792$ & \cellcolor[RGB]{204, 229, 255}$24.365$ & \cellcolor[RGB]{204, 229, 255}$9.301$ & \cellcolor[RGB]{204, 229, 255}$0.011$ & $0.007$ & $0.008$ & \cellcolor[RGB]{230, 255, 230}$36.484$ & \cellcolor[RGB]{204, 229, 255}$3.045$ & \cellcolor[RGB]{204, 229, 255}$40.657$ & \cellcolor[RGB]{204, 229, 255}$11.308$ & \cellcolor[RGB]{204, 229, 255}$0.014$ & $0.007$ & $0.010$ & \cellcolor[RGB]{230, 255, 230}$55.041$ \\
TransMIL~\cite{shao2021transmil} & \cellcolor[RGB]{204, 229, 255} & \cellcolor[RGB]{204, 229, 255} & \cellcolor[RGB]{204, 229, 255} & \cellcolor[RGB]{204, 229, 255} & $0.030$ & $0.009$ & \cellcolor[RGB]{230, 255, 230}$36.508$ & \cellcolor[RGB]{204, 229, 255} & \cellcolor[RGB]{204, 229, 255} & \cellcolor[RGB]{204, 229, 255} & \cellcolor[RGB]{204, 229, 255} & $0.030$ & $0.010$ & \cellcolor[RGB]{230, 255, 230}$55.064$ \\
DTFD-AFS~\cite{zhang2022dtfd} & \cellcolor[RGB]{204, 229, 255} & \cellcolor[RGB]{204, 229, 255} & \cellcolor[RGB]{204, 229, 255} & \cellcolor[RGB]{204, 229, 255} & $0.003$ & $0.008$ & \cellcolor[RGB]{230, 255, 230}$36.480$ & \cellcolor[RGB]{204, 229, 255} & \cellcolor[RGB]{204, 229, 255} & \cellcolor[RGB]{204, 229, 255} & \cellcolor[RGB]{204, 229, 255} & $0.003$ & $0.009$ & \cellcolor[RGB]{230, 255, 230}$55.036$ \\
DTFD-MAS~\cite{zhang2022dtfd} & \cellcolor[RGB]{204, 229, 255} & \cellcolor[RGB]{204, 229, 255} & \cellcolor[RGB]{204, 229, 255} & \cellcolor[RGB]{204, 229, 255} & $0.004$ & $0.009$ & \cellcolor[RGB]{230, 255, 230}$36.482$ & \cellcolor[RGB]{204, 229, 255} & \cellcolor[RGB]{204, 229, 255} & \cellcolor[RGB]{204, 229, 255} & \cellcolor[RGB]{204, 229, 255} & $0.004$ & $0.011$ & \cellcolor[RGB]{230, 255, 230}$55.039$ \\
S4MIL~\cite{fillioux2023structured} & \cellcolor[RGB]{204, 229, 255} & \cellcolor[RGB]{204, 229, 255} & \cellcolor[RGB]{204, 229, 255} & \cellcolor[RGB]{204, 229, 255}& $0.006$ & $0.014$ & \cellcolor[RGB]{230, 255, 230}$36.489$ & \cellcolor[RGB]{204, 229, 255} & \cellcolor[RGB]{204, 229, 255} & \cellcolor[RGB]{204, 229, 255} & \cellcolor[RGB]{204, 229, 255} & $0.007$ & $0.017$ & \cellcolor[RGB]{230, 255, 230}$55.048$ \\
MambaMIL~\cite{yang2024mambamil} & \cellcolor[RGB]{204, 229, 255} & \cellcolor[RGB]{204, 229, 255} & \cellcolor[RGB]{204, 229, 255} & \cellcolor[RGB]{204, 229, 255} & $0.056$ & $0.009$ & \cellcolor[RGB]{230, 255, 230}$36.534$ & \cellcolor[RGB]{204, 229, 255} & \cellcolor[RGB]{204, 229, 255} & \cellcolor[RGB]{204, 229, 255} & \cellcolor[RGB]{204, 229, 255} & $0.056$ & $0.011$ & \cellcolor[RGB]{230, 255, 230}$55.091$ \\
\rowcolor[RGB]{255,242,204}AHDMIL & $2.792$ & $18.298$ & $6.213$ & $0.016$ & $0.015$ & $0.028$ & $27.362$ & $3.045$ & $19.016$ & $3.959$ & $0.020$ & $0.015$ & $0.031$ & $26.086$ \\
\bottomrule
\end{tabular}
\end{sidewaystable*}

\begin{table*}[htbp]
\caption{
The contribution of each component in AHDMIL to the classification performance.
``SD", ``FT", ``SL" and ``DL" denote self-distillation, fine-tuning, single-branch LIPN, and double-branch LIPN, respectively.
Rows in light green, pink, and yellow correspond to results using only DMIN, our conference version HDMIL, and the proposed AHDMIL, respectively.
}
\setlength{\tabcolsep}{4.5pt}
\centering
\begin{tabular}{ccccc|cccccccc|cc}
\toprule
\multicolumn{2}{c}{Stage1} & \multicolumn{3}{c|}{Stage2} & \multicolumn{2}{c}{Camelyon16} & \multicolumn{2}{c}{TCGA-NSCLC} & \multicolumn{2}{c}{TCGA-BRCA} & \multicolumn{2}{c|}{TCGA-RCC} & \multicolumn{2}{c}{Average} \\
\cmidrule(r){1-2} \cmidrule(r){3-5} \cmidrule(r){6-7} \cmidrule(r){8-9} \cmidrule(r){10-11} \cmidrule(r){12-13} \cmidrule(r){14-15}
CKA & SD & FT & SL & DL & AUC & Acc & AUC & Acc & AUC & Acc & AUC & Acc & AUC & Acc \\
\midrule
\ding{55} & \ding{55} & \ding{55} & \ding{55} & \ding{55} & $87.51$ & $82.56$ & $95.59$ & $88.01$ & $90.22$ & $88.27$ & $98.76$ & $92.20$ & $93.02$ & $87.76$ \\
\ding{51} & \ding{55} & \ding{55} & \ding{55} & \ding{55} & $91.21$ & $85.89$ & $96.16$ & $89.73$ & $90.02$ & $88.47$ & $98.78$ & $93.33$ & $94.04$ & $89.36$ \\
\rowcolor[RGB]{204,255,242} \ding{51} & \ding{51} & \ding{55} & \ding{55} & \ding{55} & $93.17$ & $88.92$ & $96.47$ & $89.75$ & $90.43$ & $88.68$ & $98.74$ & $92.66$ & $94.70$ & $90.00$  \\
\rowcolor[RGB]{242,204,255} \ding{51} & \ding{51} & \ding{55} & \ding{51} & \ding{55} & $90.88$ & $88.61$ & $96.35$ & $89.78$ & $90.45$ & $88.27$ & $98.72$ & $91.38$ & $94.10$  & $89.51$ \\
\ding{51} & \ding{51} & \ding{51} & \ding{51} & \ding{55} & $90.16$ & $86.67$ & $96.59$ & $89.48$ & $89.88$ & $88.57$ & $98.82$ & $93.09$ & $93.86$ & $89.45$ \\
\rowcolor[RGB]{255,242,204} \ding{51} & \ding{51} & \ding{51} & \ding{55} & \ding{51} & $91.96$ & $89.92$ & $96.64$ & $90.72$ & $90.93$ & $88.87$ & $99.02$ & $93.78$ & $94.64$ & $90.82$ \\
\bottomrule
\end{tabular}
\label{tab:each_component}
\end{table*}

\begin{figure*}[!t]
    \centering
    \includegraphics[width=1.0\textwidth]{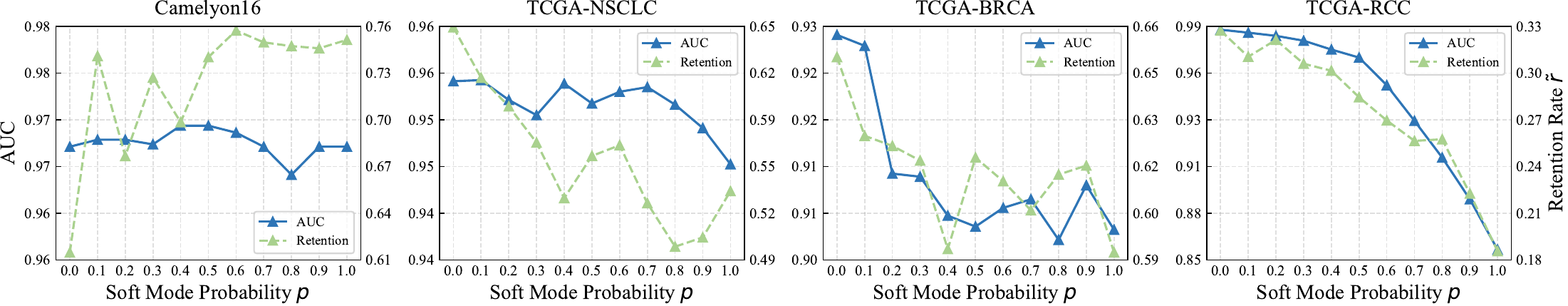}
    \caption{10-fold average AUC and learned instance retention rate $\tilde{r}$ under varying soft mode probability $p$.}
    \label{fig:soft_probability}
\end{figure*}

\subsection{Ablation Study}
\label{sec:Comprehensive Analysis}
We conduct ablation studies to gain deeper insights into the internal mechanisms and design rationale of AHDMIL.
Specifically, we analyze the contribution of each module to classification performance, study the effect of varying the soft mode probability $p$ in CRD, and examine factors influencing the CKA classifier.

For all experiments involving the inclusion or removal of specific modules, we report results on the 10-fold test sets to reflect generalization performance.
For analyses related to internal design choices or hyperparameter tuning (e.g., soft mode probability $p$ or CKA classifier settings), we report results on the 10-fold validation sets, as these experiments are intended to guide model design rather than evaluate final performance.

\subsubsection{Contribution of Ecah Module}
Table~\ref{tab:each_component} summarizes the contribution of each component in AHDMIL to the final classification performance.
The first row shows the standard CLAM-MB setup, serving as the baseline for our DMIN module.
Replacing the original classifier in CLAM-MB with the proposed CKA classifier yields a clear performance improvement.
Incorporating self-distillation into DMIN further enhances performance, particularly on datasets like Camelyon16, where tumor regions occupy a small portion of WSIs.
When knowledge is distilled from a fixed DMIN into a single-branch LIPN (corresponding to HDMIL), limited validation data hinders unbiased model selection and leads to performance degradation~\cite{dong2025fast}.
To mitigate this, we introduce an asymmetric distillation mechanism to fine-tune DMIN, better aligning it with the input distribution after instance selection.
This approach improves certain metrics on some datasets but achieves overall performance comparable to HDMIL.
Finally, adopting the dual-branch LIPN enhances model robustness via heterogeneity between branches, helping to alleviate the negative impact of limited validation data on unbiased model selection.
This strategy even enables the student model to outperform its teacher (DMIN) in some cases.
Overall, this stepwise ablation study highlights how each component progressively contributes to AHDMIL’s effectiveness, validating our modular design and demonstrating significant gains in both robustness and classification performance.

\subsubsection{Effect of Soft Mode Probability $p$}
Figure~\ref{fig:soft_probability} illustrates how varying the soft mode probability $p$ from 0.0 to 1.0 in cross-resolution distillation affects both classification performance and instance retention rates $\tilde{r}$.
On Camelyon16, the classification AUC peaks at moderate $p$ values, with $\tilde{r}$ initially increasing and then declining as $p$ grows.
For the TCGA datasets (TCGA-BRCA, TCGA-NSCLC, and TCGA-RCC), the highest AUCs are observed at lower $p$ values, with both AUC and $\tilde{r}$ generally decreasing as $p$ increases.
These trends indicate a close alignment between classification performance and instance retention, since an excessively low $\tilde{r}$ often reflects a high risk of discarding informative instances, ultimately leading to degraded performance.
Moreover, these differences in trends across datasets stem from their distinct characteristics, such as sample size and complexity, which influence how the model leverages soft mode information during distillation.
For Camelyon16, the relatively small training set makes it difficult for the model to learn effectively from hard mode distillation alone, so moderate soft mode guidance provides necessary fine-grained cues.
However, excessive soft mode distillation (i.e., high $p$) may introduce noise that may overwhelm discriminative signals.
In contrast, the TCGA datasets have ample training samples, making hard mode distillation sufficient. 
In such cases, excessive soft mode information can dilute critical decision boundaries, harming performance.

\begin{figure*}[!t]
    \centering
    \includegraphics[width=1.0\textwidth]{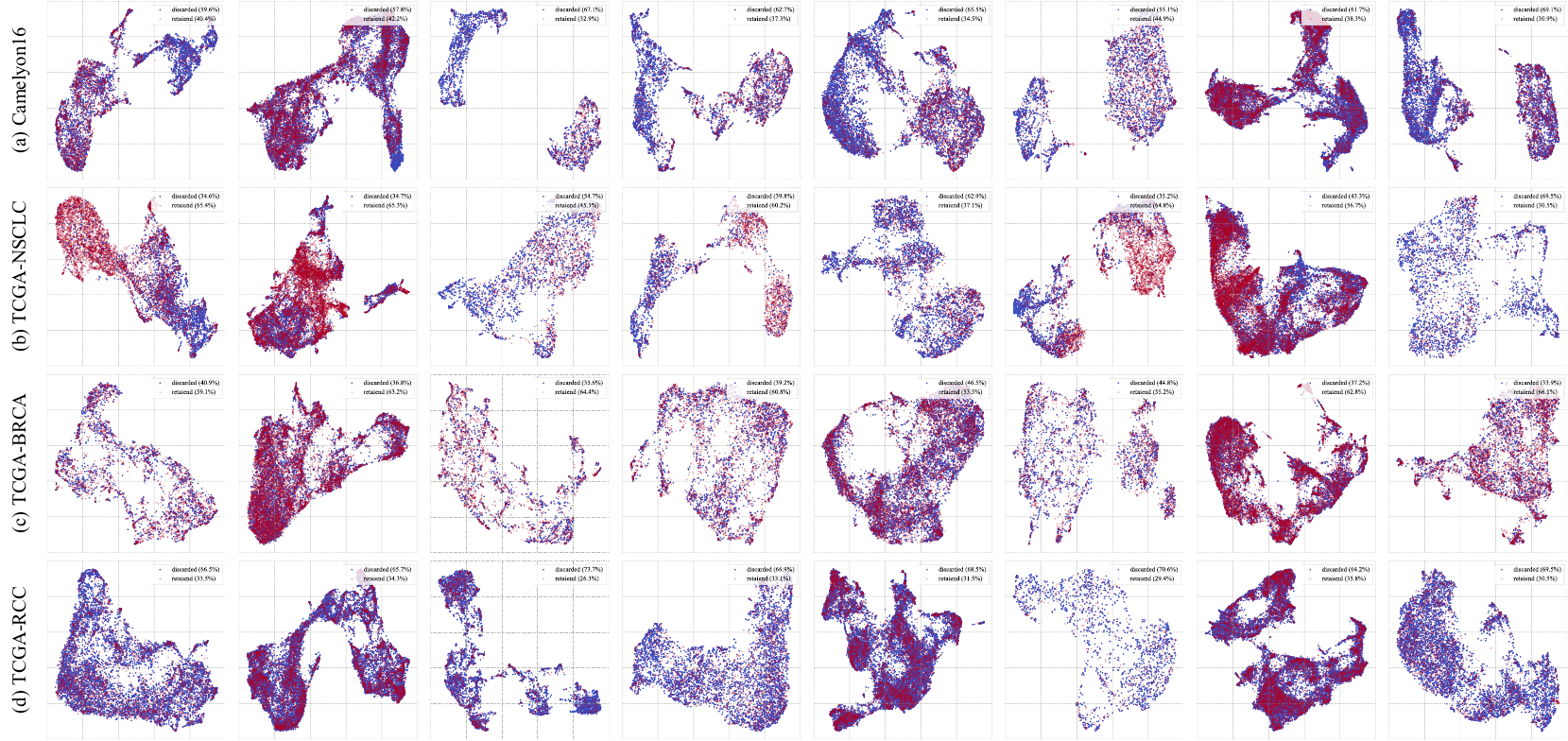}
    \caption{
    UMAP visualization of instance embeddings. Instances retained by AHDMIL are marked in red, while discarded ones are marked in blue.
    }
    \label{fig:umap}
\end{figure*}

\subsubsection{Factors Influencing the CKA Classifier}
\begin{table}[!h]
\caption{Ablation study on different CKA classifier design choices. Reported results are 10-fold validation performance on the Camelyon16 dataset. The final configuration adopted in our main method is highlighted in yellow for each group.}
\setlength{\tabcolsep}{3.8pt}
\centering
\begin{tabular}{llccc}
\toprule
Group & Setting & Params (M) & AUC & Acc \\
\midrule
\multirow{3}{*}{Initialization} & Normal~\cite{krizhevsky2012imagenet} & $0.83$ & $96.42$ & $90.38$ \\
~ & Kaiming~\cite{he2015delving} & $0.83$ & $94.12$ & $89.23$ \\
~ & \cellcolor[RGB]{255,242,204}Xaiver~\cite{glorot2010understanding} & \cellcolor[RGB]{255,242,204}$0.83$ & \cellcolor[RGB]{255,242,204}$97.15$ & \cellcolor[RGB]{255,242,204}$93.85$ \\

\midrule
\multirow{3}{*}{Position} & Projection & $7.08$ & $92.79$ & $86.54$ \\ 
~ & Attention & $3.94$ & $85.03$ & $77.31$ \\ 
~ & \cellcolor[RGB]{255,242,204}Classifier & \cellcolor[RGB]{255,242,204}$0.83$ & \cellcolor[RGB]{255,242,204}$97.15$ & \cellcolor[RGB]{255,242,204}$93.85$ \\ 

\midrule
\multirow{4}{*}{Head} & FC & $0.79$ & $94.67$ & $91.54$ \\
~ & MLP & $1.84$ & $94.97$ & $92.69$ \\
~ & KA~\cite{liu2024kan} & $0.83$ & $96.42$ & $91.16$ \\
~ & \cellcolor[RGB]{255,242,204}CKA & \cellcolor[RGB]{255,242,204}$0.83$ & \cellcolor[RGB]{255,242,204}$97.15$ & \cellcolor[RGB]{255,242,204}$93.85$ \\

\midrule
\multirow{4}{*}{Degree} & $K=4$ & $0.80$ & $94.67$ & $89.62$ \\
~ & $K=8$ & $0.82$ & $94.61$ & $90.38$ \\
~ & \cellcolor[RGB]{255,242,204}$K=12$ & \cellcolor[RGB]{255,242,204}$0.83$ & \cellcolor[RGB]{255,242,204}$97.15$ & \cellcolor[RGB]{255,242,204}$93.85$ \\
~ & $K=16$ & $0.84$ & $96.18$ & $90.39$ \\
\bottomrule
\end{tabular}
\label{tab:CKA}
\end{table}

As shown in Table~\ref{tab:CKA}, we analyze the proposed CKA classifiers from four perspectives: 1) the impact of different initialization for the learnable coefficients; 2) the impact of employing the CKA layer at different positions; 3) performance comparison with other classification heads; and 4) the impact of different degrees $K$. 
To eliminate the impact of other factors, all experiments here only utilize the teacher branch of DMIN.
This ablation study reveals several observations:
\begin{itemize}
    \item Appropriate initialization matters. It can be seen that when Normal~\cite{krizhevsky2012imagenet} or Kaiming~\cite{he2015delving} is used instead of Xaiver~\cite{glorot2010understanding} initialization, the classification performance is significantly reduced, especially in terms of accuracy. 
    \item When using CKA layers in the projection or attention module, the number of trainable parameters increases dramatically, accompanied by a significant drop in performance. 
    \item When compared with other classifiers such as the FC layer, two-layer MLP, and KA~\cite{liu2024kan} layer, CKA demonstrates superior classification performance.
    Although FC, MLP, and KA can sometimes achieve similar performance to CKA in certain folds, there tends to be a larger performance gap in other folds.
    Thus, CKA is a more powerful and robust classifier.
    \item When the Chebyshev polynomial degree $K$ changes from 4 to 16, the number of parameters of the entire DMIN does not change much. Nevertheless, there is a noticeable disparity in classification performance, with the best outcome achieved at $K=12$. Further increasing the order does not lead to better improvements in classification performance, probably due to the limited training data.
\end{itemize}

\subsection{Visualization Analysis}
To gain visual insights into how AHDMIL filters out irrelevant instances, we perform UMAP~\cite{mcinnes2018umap} visualizations for each WSI, labeling retained instances in red and discarded ones in blue, as shown in Figure~\ref{fig:umap}.
For each dataset, we select the model from the fold with the lowest average instance retention ratio and randomly visualize 8 WSIs.
Notably, the distributions of retained and discarded instances often show substantial overlap in the embedded space, suggesting that AHDMIL does not filter instances based on their separability in feature space.
Instead, the model aims to reduce redundancy by removing repetitive instances—even those sharing similar semantic content with the retained ones.
This selective filtering facilitates a compact yet diverse representation of relevant patterns within each bag, enhancing generalization without sacrificing critical information.

\section{Conclusion}\label{sec:conclusion}
In this paper, we propose AHDMIL, a novel framework for accelerating WSI classification while maintaining high accuracy.  
Through asymmetric hierarchical distillation, AHDMIL efficiently filters out irrelevant patches in WSIs, significantly reducing preprocessing time.  
Extensive experiments on four public datasets spanning three different organ sites demonstrate that AHDMIL outperforms current state-of-the-art methods in both classification performance and inference speed.

In future work, we plan to further improve inference efficiency by optimizing the feature extraction process for retained instances, which remains a major computational component.  
We also aim to enhance instance selection by exploring sparse attention mechanisms or alternative strategies beyond standard attention.  
These directions build naturally on our current AHDMIL, which already demonstrates strong improvements in both inference speed and classification performance.

\section*{Declarations}
\begin{itemize}
\item Funding: This work was supported in part by the National Natural Science Foundation of China under 62031023 \& 62331011, and in part by the Shenzhen Science and Technology Project under GXWD20220818170353009.
\item Conflict of interest: The authors declare that they have no conflict of interest.
\item Ethics approval and consent to participate: Not applicable.
\item Consent for publication: Yes.
\item Data availability: All datasets used in this work are publicly available. The Camelyon16 dataset is available at https://camelyon16.grand-challenge.org/Download/. The TCGA-NSCLC, TCGA-BRCA, and TCGA-RCC datasets are available at https://portal.gdc.cancer.gov/.
\item Materials availability: Not applicable.
\item Code availability: The code is available at https://github.com/JiuyangDong/AHDMIL.
\item Author contribution: Jiuyang Dong and Jiahan Li contributed to the conception and design of the study. 
Jiuyang Dong performed the experiments and wrote the initial manuscript. 
Jiahan Li conducted the data analysis and visualization. 
Yongbing Zhang, Junjun Jiang, and Kui Jiang supervised the project, provided critical feedback, and revised the manuscript. 
All authors read and approved the final version of the manuscript.

\end{itemize}


\end{document}